\documentclass[11pt]{article}

\usepackage[]{EMNLP2023}

\usepackage{times}
\usepackage{latexsym}

\usepackage[T1]{fontenc}

\usepackage[utf8]{inputenc}

\usepackage{microtype}

\usepackage{inconsolata}
\usepackage{times}
\usepackage{epsfig}
\usepackage{graphicx}
\usepackage{amsmath}
\usepackage{amssymb}
\usepackage{booktabs}
\usepackage{multirow}
\usepackage{tabularx}
\usepackage{pgfplots}
\usepackage{xspace}
\usepackage{xcolor}
\usepackage{colortbl}
\usepackage{subcaption}
\usepackage{enumitem}
\usepackage{comment}
\usepackage{wasysym}
\usepackage{amssymb}
\usepackage{pifont}
\newcommand{\cmark}{\ding{51}}%
\newcommand{\xmark}{\ding{55}}%

\makeatletter
\@namedef{ver@everyshi.sty}{}
\makeatother
\usepackage{tikz,pgfplots}
\pgfplotsset{compat=1.14}

\usepackage{etoolbox}
\definecolor{Azul}{rgb}{0.16, 0.32, 0.75}
\usepackage{soul}

\title{Can Pre-trained Vision and Language Models Answer \\ Visual Information-Seeking Questions?}

\author{
{Yang Chen}$^{\spadesuit\heartsuit}$\thanks{\quad Work done when interned at Google}\quad {Hexiang Hu}$^{\spadesuit}$\quad {Yi Luan}$^{\spadesuit}$\quad {Haitian Sun}$^{\spadesuit}$ \quad {Soravit Changpinyo}$^{\clubsuit}$ \quad \\
{\textbf{Alan Ritter}}$^{\heartsuit}$ \quad {\textbf{Ming-Wei Chang}}$^{\spadesuit}$ \\
$^{\spadesuit}$\xspace {Google Deepmind} \quad
$^{\clubsuit}$\xspace {Google Research} \quad$^{\heartsuit}$\xspace {Georgia Institute of Technology}
}

\begin{document}

\newcommand{\backup}[1]{{}}
\newcommand{\nlp}[1]{\texttt{\small #1}}
\newcommand{\snlp}[1]{\texttt{\small #1}}
\newcommand{\entity}[1]{\texttt{\textsc{\small #1}}}

\newcommand{\frank}[1]{{\color{brown}Frank: #1}\xspace}
\newcommand{\yi}[1]{{\color{purple}Yi: #1}\xspace}
\newcommand{\mw}[1]{{\color{blue}Ming-Wei: #1}\xspace}
\newcommand{\yc}[1]{{\color{red}Yang: #1}\xspace}
\newcommand{\ar}[1]{{\color{green}Alan: #1}\xspace}
\newcommand{\beer}[1]{{\color{orange}Beer: #1}\xspace}

\newcommand{\ie}{\textit{i.e}}
\newcommand{\eg}{{\small e.g.}}
\newcommand\mypara[1]{\vspace{1mm}\noindent\textbf{#1}}

\newcommand{\ourdataset}{\textsc{InfoSeek}\xspace}
\newcommand{\infoseek}{\ourdataset}
\newcommand{\palift}{PaLI\xspace}
\newcommand{\ofaft}{OFA\xspace}

\newcommand{\qonly}{PaLM (Q-only)\xspace}

\newcommand{\seen}[0]{{\textsc{Seen}}\xspace}
\newcommand{\unseen}[0]{{\textsc{Unseen}}\xspace}
\newcommand{\nokb}[0]{{\textbf{No-KB}}\xspace}
\newcommand{\withkb}[0]{{\textbf{With-KB}}\xspace}
\definecolor{citecolor}{RGB}{30,130,255}

\newcommand{\pd}{\hphantom{.}}
\newcommand{\pz}{\hphantom{0}}
\newcommand{\pzz}{\hphantom{00}}

\definecolor{lightgray}{gray}{0.9}
\definecolor{beige}{rgb}{0.96, 0.96, 0.86}
\definecolor{blanchedalmond}{rgb}{1.0, 0.92, 0.8}

\newcommand{\custompara}[1]{{\vspace{1mm}\noindent\textbf{#1}\xspace}}
\newcommand{\entgen}[0]{\textsc{EntGen}\space}
\newcommand{\entret}[0]{\textsc{EntRet}\space}

\newcommand*{\ClipSep}{0.1cm}%




\maketitle
\begin{abstract}
    Pre-trained vision and language
models~\cite{chen2022pali,chen2023pali,dai2023instructblip,li2023blip} have demonstrated state-of-the-art capabilities over existing tasks involving images and texts, including visual question answering. However, it remains unclear whether these models possess the capability to answer questions that are not only querying visual content but knowledge-intensive and information-seeking. In this study, we introduce \infoseek\footnote{Our dataset is available at \url{https://open-vision-language.github.io/infoseek/}.}, a visual question answering dataset tailored for information-seeking questions that cannot be answered with only common sense knowledge. 
Using \infoseek, we analyze various pre-trained visual question answering models and gain insights into their characteristics. Our findings reveal that state-of-the-art pre-trained multi-modal models (\eg, PaLI-X, BLIP2, etc.) face challenges in answering visual information-seeking questions, but fine-tuning on the \ourdataset dataset elicits models to use fine-grained knowledge that was learned during their pre-training.
Furthermore, we show that accurate visual entity recognition can be used to improve performance on \ourdataset by retrieving relevant documents, showing a significant space for improvement.
\end{abstract}

\section{Introduction}
\label{sec:intro}
\begin{figure}[t]
    \centering
    \includegraphics[width=0.49\textwidth]{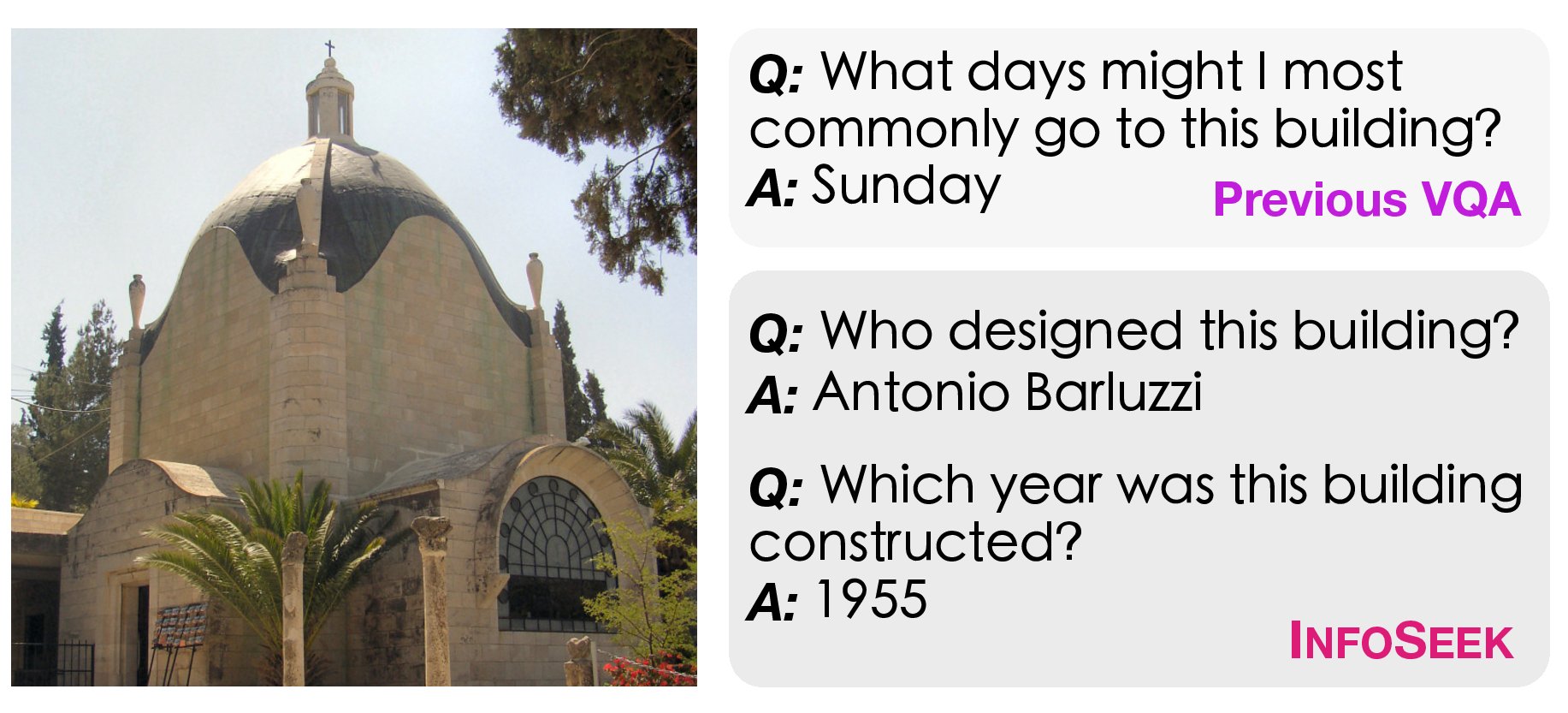}
    \caption{
        While 70.8\% of OK-VQA questions can be answered by average adults without using a search engine, \ourdataset poses challenges to query fine-grained information about the visual entity (\eg, \nlp{Dominus Flevit Church}), resulting in a sharp drop to 4.4\% (\S\ref{sec:formulation}).
    }
    \label{fig:intro_example}
\end{figure}

The acquisition of knowledge occurs in the pre-training of large language models~\cite{brown2020language,chowdhery2022palm}, demonstrated as their emergent ability to answer information-seeking questions in the open-world, where the questioner does not have easy access to the information. 
While prior works have analyzed models' capabilities to answer {\em textual} information-seeking (or info-seeking) questions, much less is known for {\em visual} info-seeking questions. For example, after taking a picture of the specific church in Figure~\ref{fig:intro_example}, a person might want to know the date of construction, or who decorated the interior of the church. Although the entity is presented in the image (the specific church), the relevant knowledge (\eg, the date) is not.
Given recent advances on pre-trained visual and language models~\cite{alayrac2022flamingo, chen2022pali, li2023blip}, {\em {do these models also understand how to answer visual information-seeking questions?}}

To study this research question, a visual question answering (VQA) dataset focusing on info-seeking questions is inevitably required. However, not all VQA datasets meet this criterion. For example, 
by design, the majority of questions in 
datasets such as VQA v2~\cite{balanced_vqa_v2} focus on visual attributes and object detection that does not require information beyond the image to answer.
While models capable of answering these types of questions have the potential to aid visually impaired individuals~\cite{gurari2018vizwiz}, there is a broader class of {\em info-seeking} questions that cannot be easily answered by sighted adults. Handling such questions (\eg, \nlp{When was this building constructed?~1955}) is critical as they come closer to the natural distribution of human questions. 

In this paper, we present \infoseek, a natural VQA dataset that focuses on visual {info-seeking} questions. 
Different from previous VQA datasets, the testing subset of \ourdataset is collected in multiple stages from human annotators to evaluate VQA where the question can not be answered from only the visual content (see a comparison of datasets in \S~\ref{sec:formulation}).
In addition to this manually curated test set, which enables realistic evaluation of info-seeking VQA, we also join annotations from a recent visual entity recognition dataset~\cite{hu2023opendomain} with the Wikidata database~\cite{vrandevcic2014wikidata}, and employ human annotators to write templates to semi-automatically generate a large corpus of visual info-seeking QA pairs. 
Over 1 million $\{\text{image, question, answer}\}$ triplets are generated to support fine-tuning multimodal models for info-seeking VQA. We split data to ensure memorizing knowledge during fine-tuning is useless --- models either have to learn to use knowledge learned during pre-training or learn to retrieve knowledge from an external knowledge base.

Using \infoseek, we analyze the ability of state-of-the-art models to answer visual info-seeking questions. We found pre-trained vision-language models, such as models pre-trained end-to-end (\eg, PaLI-X by~\citeauthor{chen2023pali}), and models pre-trained with frozen LLM (\eg, BLIP2 by~\citeauthor{li2023blip}), both struggle to answer info-seeking questions in zero-shot, though BLIP2 outperforms PaLI-X by a margin. 
Surprisingly, after fine-tuning on our (large, semi-automatically curated) training set, PaLI-X yields a significant improvement and outperforms the fine-tuned BLIP2 models on queries that are unseen during fine-tuning. This suggests that while pre-trained PaLI-X has a significant amount of knowledge, it requires a small amount of fine-tuning data to fully awaken its capabilities.
Furthermore, we show that \infoseek~fine-tuned models can even generalize to questions and entity types completely unseen during fine-tuning (\eg, art \& fashion).

When incorporating a visual entity recognition component, and conditioning models on the Wikipedia articles of the relevant entities, we show that models accessing such a knowledge base (With-KB) perform better overall than those that rely on knowledge learned during pre-training. However, end-to-end (No-KB) models were found better on certain classes of questions that require coarse-grained answers ({\em ``Which continent is this building located on?''}), even on tail entities. Our experiment (\S \ref{subsec:results_withkb}) further suggests that improving visual entity recognition can drastically increase model's capability in answering visual info-seeking questions (from 18\% to 45.6\%), indicating a promising direction for future development.

\section{The Need for a New Visual Information-seeking Benchmark}
\label{sec:formulation}

\begin{table}[t]
\tabcolsep 8pt
\small
\begin{tabular}{@{}l@{}ccc@{\;}}
\toprule
Dataset & {\small OK-VQA} & {\small ViQuAE} & {\small \ourdataset}\cellcolor{beige} \\
\midrule
\qonly & 23.8 & {\bf 31.5} & \pz5.6\cellcolor{beige}\\
Current SotA & {66.1} & 22.1 & {18.2}\cellcolor{beige}\\
\midrule
Require Knowledge$^\dagger$ & 29.2\% & 95.2\% & 95.6\%\cellcolor{beige} \\
\bottomrule
\end{tabular}
{\scriptsize $\dagger:$\% of questions that require knowledge to answer.} \\
{\scriptsize PaLM (Q-only): a question-only baseline using PaLM.}
\caption{
    Comparison of {\infoseek} and prior KI-VQA benchmarks. Performances reported in VQA score. 
}
\label{tab:q_only}
\end{table}

While there have been plenty of knowledge-intensive VQA (KI-VQA) benchmarks, we show that none of these meet the criteria to effectively evaluate info-seeking VQA.
Early efforts in this area, such as KBQA~\cite{wang2015kbqa} and FVQA~\cite{wang2017fvqa}, were based on domain-specific knowledge graphs, while recent datasets like OK-VQA~\cite{marino2019ok} and its variants such as S3VQA~\cite{jain2021select} and A-OKVQA~\cite{schwenk2022aokvqa} have improved upon this foundation by incorporating an open-domain approach and highlighting common-sense knowledge. Among the existing benchmarks, K-VQA~\cite{kvqa} and ViQuAE~\cite{viquae} are the most relevant, but they have severe limitations in their question generation process, as discussed below.

\custompara{Information Seeking Intent.}
The evaluation of models' ability to answer info-seeking questions requires fine-grained knowledge, which a person is unlikely to know off the top of their head.
However, we found that 70.8\% of OK-VQA questions\footnote{Studied with human on 500 random OK-VQA questions (see Appendix~\ref{appendix:okvqa})} can be answered without the need to use a search engine, indicating the dataset primarily focuses on knowledge that is commonly known to people. 
Most OK-VQA questions are regarding coarse-grained knowledge that many people already know: \nlp{What days might I most commonly go to this building?~Sunday}. One only needs to know the building type (\eg, \nlp{Church}) rather than the specific building (\eg, \nlp{Dominus~Flevit~Church}).
This makes it unsuitable for evaluating pre-trained models on long-tailed knowledge, where these models have shown weaknesses~\cite{kandpal2022large}.

\custompara{Reliance on Visual Understanding.}
In contrast to OK-VQA, the ViQuAE dataset aims to test fine-grained knowledge of visual entities by pairing questions from TriviaQA~\cite{joshi-etal-2017-triviaqa} with images.
However, a significant portion of the ViQuAE questions {(\eg, \snlp{"Who betrayed him for 30 pieces of silver?"})} can be answered without looking at the images, as the questions often reveal sufficient information to determine the answer. 
To quantify this observation, we present questions from the evaluation set to a large language model, PaLM (540B)~\cite{chowdhery2022palm}.
Results on the ViQuAE test set are shown in Table~\ref{tab:q_only}. Surprisingly, we find that PaLM can read questions and generate answers with 31.5\% accuracy, outperforming the SOTA retrieval-based model~\cite{viquae} (which has access to the image) on this dataset by 9.4\%.
Although PaLM is a much larger model, this experiment illustrates that it is possible to achieve very good performance on ViQuAE without using information from the image.

\custompara{Entity Coverage.}
Current VQA datasets often cover a limited number of visual entity categories.
For example, K-VQA only focuses on human subjects, while over 43\% of questions in ViQuAE revolve around human entities (see Table~\ref{tab:datasets}).
Such limitations hinder the evaluation of a model's knowledge across various entity categories and may result in reduced task complexity, as the evaluation may be limited to mere facial recognition.

To address these limitations, we present \ourdataset (\S~\ref{sec:dataset}), a new benchmark for pre-trained multi-modal models on visual info-seeking questions.
Our work builds on top of a visual entity recognition dataset,
OVEN~\cite{hu2023opendomain}, which is designed to answer questions related to the identification of visual entities.
We take visual info-seeking a step further by benchmarking info-seeking questions about visual entities, which allows us to test the pre-training knowledge of models beyond simply recognizing an entity.

\begin{figure*}[!htbp]
    \centering
    \vspace{-5mm}
    \begin{tabular}{c@{\quad}c}
        \tabcolsep 0pt
        \includegraphics[height=3.3cm]{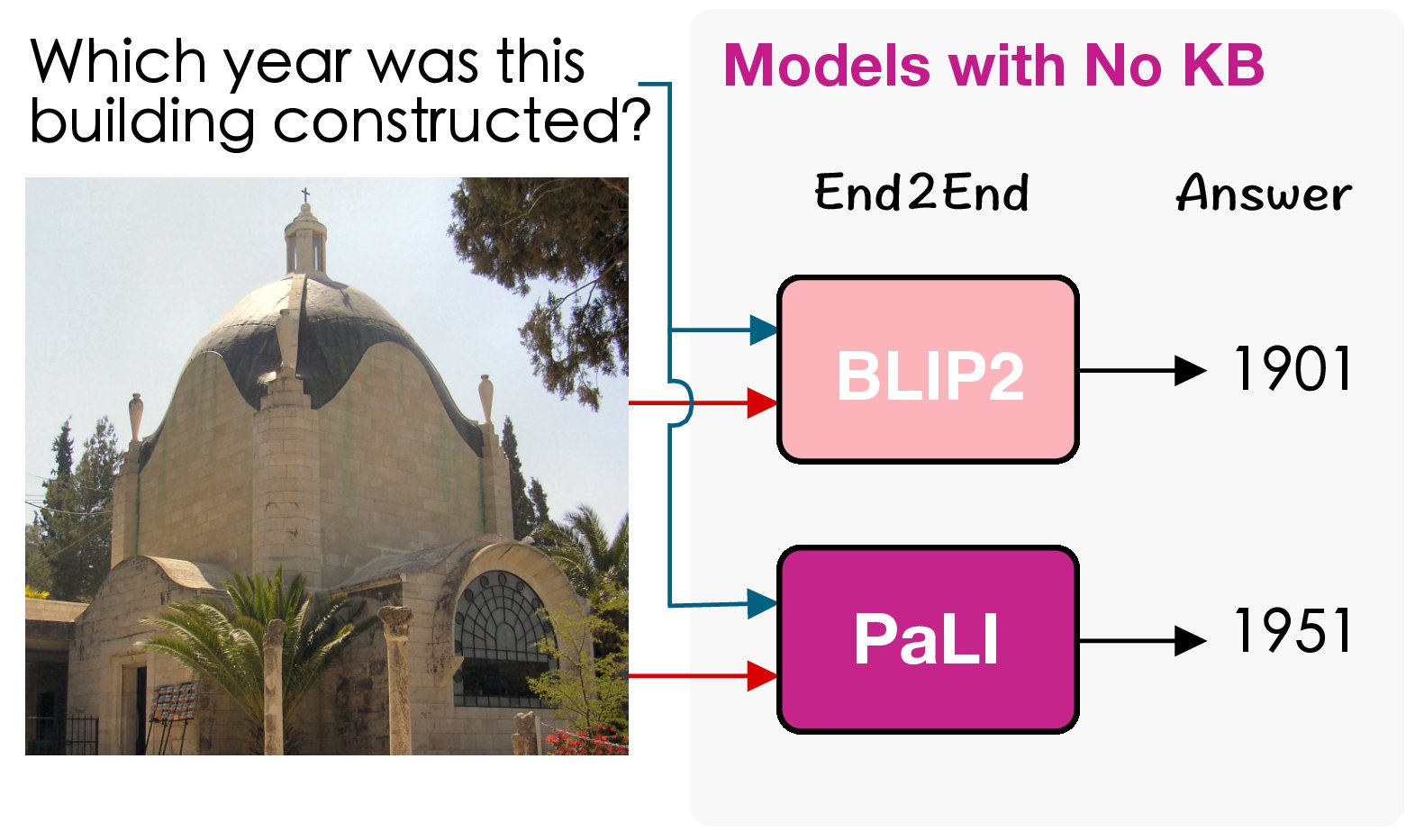} & \includegraphics[height=3.3cm]{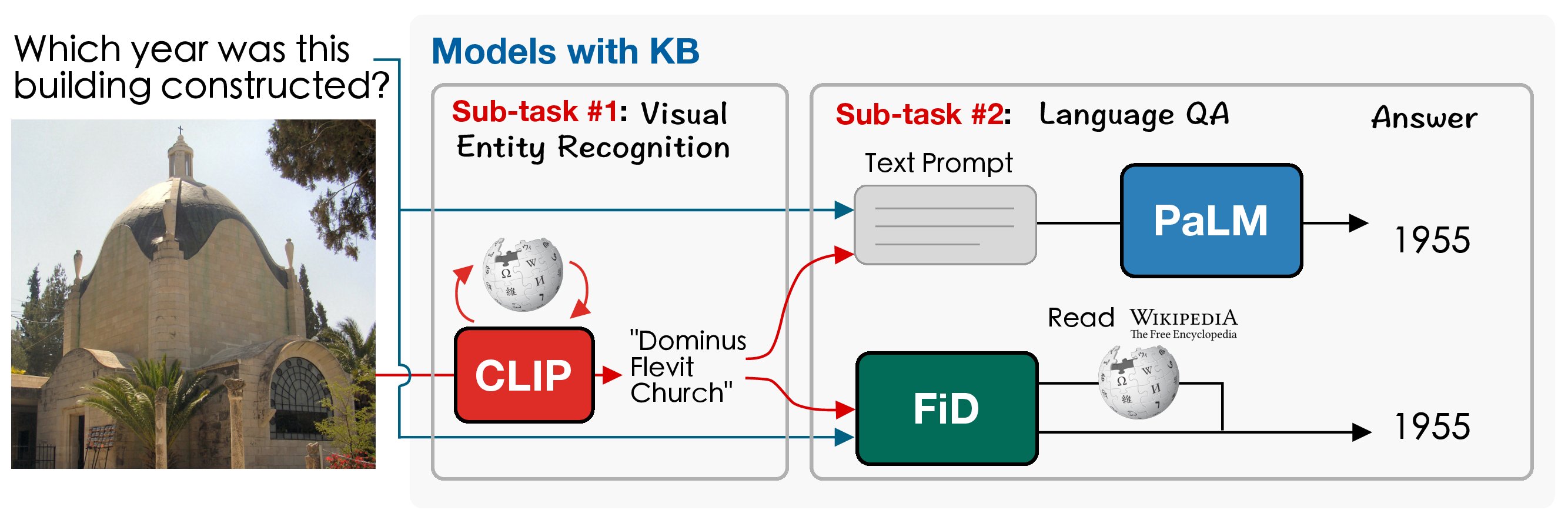} \\
        {\footnotesize (a) Models with \textbf{No KB} access} & {\footnotesize (b) Models \textbf{With KB} {(\small \textit{Knowledge-base})} information}
    \end{tabular}
    \caption{
    \textbf{Visual info-seeking models} under the proposed \textbf{No KB} and \textbf{With KB} protocols. 
    (a) End-to-end VQA models (such as PaLI~\cite{chen2022pali} or BLIP2~\cite{li2023blip}) that directly predict the answer from looking at the image and question; 
    (b) Pipeline systems with access to a knowledge base (\eg Wikipedia), with the option to link the queried subject to the Wikipedia use CLIP~\cite{radford2021clip} and perform textual question-answering using PaLM~\citep{chowdhery2022palm} or Fusion-in Decoder (FiD)~\cite{izacard2020fid}. 
    }
    \label{fig:models}
\end{figure*}
\section{{\infoseek}: A VQA Benchmark of Visual Information-seeking Questions}
\label{sec:dataset}
The \ourdataset dataset consists of two components, (1) \ourdataset$_{\text{Human}}$: a collection of human-written visual info-seeking questions (8.9K) to simulate information seeking intent (see \S~\ref{sec:dataset_human}); and (2) \ourdataset$_{\text{Wikidata}}$: an automated dataset (1.3M) to cover diverse entities for large-scale training and evaluation purposes (see \S~\ref{sec:dataset_wiki}).
We split the dataset to ensure memorizing the training set is useless, thereby emphasizing the importance of pre-training to acquire knowledge (see \S~\ref{sec:split}). Due to space limitations, we summarize the key essence in this section and defer details to the Appendix.

\begin{table}[t]
\small
\tabcolsep 4pt
\begin{tabular}{@{}l@{}cccc@{}}
\toprule
Dataset &  \# \{$\mathsf{I},\mathsf{Q},\mathsf{A}$\} & Len of $\mathsf{Q}/\mathsf{A}$ & \# Entity & \# Ent. type\\
\midrule
{\small OK-VQA} & 14K & 8.1/1.3 & - & -$^\star$ \\
{\small K-VQA} & 183K & 10.1/1.6 & 18,880 & 1$^\dagger$\\
{\small ViQuAE} &  3.6K & 12.4/1.7 & 2,397 & 980\\
\midrule
\multicolumn{3}{@{\;}l}{\ourdataset}\\
{\xspace\xspace}- Wikidata & 1.35M &  8.9/1.5 & 11,481 & 2,739\\
{\xspace\xspace}- Human & 8.9K  & 7.8/2.3 & \pz\pz806 & \pz\pz527\\
\bottomrule
\end{tabular}
{\scriptsize $\star$: OK-VQA does not specify visual entities.} \\
{\scriptsize $\dagger$: K-VQA only covers entities from the human category.}
\caption{
{Statistics of {\infoseek} \& KI-VQA datasets}.
}
\label{tab:datasets}
\end{table}

\custompara{Image Sources for Diverse Entity Coverage.}
We sourced images from 9 image classification and retrieval datasets used in~\citeauthor{hu2023opendomain}, including landmarks (17\%), animals (13\%), food (5\%), aircraft (3\%), etc. 
We utilize their annotation, which links visual entities to their corresponding Wikipedia articles, to construct our \ourdataset dataset.

\subsection{$\textbf{\ourdataset}_{\text{Human}}$: Natural Info-Seeking VQA Data Annotated by Humans}
\label{sec:dataset_human}

To ensure \ourdataset questions \textit{rely on visual understanding} and prevent models from taking shortcuts in the question without using the image, we employ a two-stage annotation approach inspired by TyDiQA~\cite{clark2020tydi}. 
This makes it unlikely questioners will have prior knowledge of the answer like SQuAD~\cite{rajpurkar2016squad}, ensuring questions with \textit{info-seeking intents}~\cite{lee-etal-2019-latent}.

\custompara{Question Writing.}
Annotators are asked to write 3-5 questions about a visual entity based on their own curiosity and information needs. To aid the question-writing process, they are prompted with visual entity images, a short description (15 words) about the entity, and a list of Wikipedia section titles. This ensures that the questions reflect a genuine interest in learning about important aspects of the entity without seeing the answer. A set of annotation rules is employed to prevent trivial questions, such as questions about visual attributes.

\custompara{Answer Labeling.} 
For each entity, we randomly assign collected questions to a different group of annotators to label answers from Wikipedia.
Annotators were shown the Wikipedia article of the entity and asked to find a concise answer to the question: a text span that is as short as possible while still forming a satisfactory answer. 
In addition, annotators categorize questions into three types: \textsc{Time} (e.g., year), \textsc{Numerical} (e.g., height)
and \textsc{String} (e.g., location). 

Finally, we construct $\{\text{image, question, answer}\}$ (IQA) triples by assigning images for the annotated QA pair of a visual entity, followed by human verification and clarification of the questions if multiple objects are presented in the image. 
Following TyDiQA~\cite{clark2020tydi}, we measure the \textit{correctness} of annotations and take the high accuracy (95\%) as evidence that the quality of the dataset is reliable for evaluating visual info-seeking models.

\subsection{$\textbf{\ourdataset}_{\text{\tiny Wikidata}}$: 1 Million Automated VQA Data from Wikipedia}
\label{sec:dataset_wiki}
\begin{table*}[!htbp]
\vspace{-2mm}
\centering
\small
\tabcolsep 12pt
\begin{tabular}{@{\;}l@{\;}c c c c c@{\;}}
\toprule
Eval Protocol & Training/Validation  &  Testing & Methods &  Example Models & Knowledge Base\\
\midrule
\nokb & \{$\mathsf{I}, \mathsf{Q}, \mathsf{A}$\}  & \{$\mathsf{I}, \mathsf{Q}$\} &End-to-end Model & PaLI, BLIP2 & - \\
\withkb & \{$\mathsf{I}, \mathsf{Q}, \mathsf{A}, \mathsf{E}$\}  & \{$\mathsf{I}, \mathsf{Q}$\} & Pipeline System & CLIP$ \to$ PaLM / FiD & Wikipedia \\
\bottomrule
\end{tabular}
\caption{\textbf{Two evaluation protocols of \infoseek.} The key difference is whether auxiliary data for visual entity recognition and knowledge base is available at training.  $\mathsf{I}$: image, $\mathsf{Q}$: question, $\mathsf{A}$: answer, $\mathsf{E}$: queried visual entity. 
}
\label{tab:eval_protocols}
\end{table*}
Human annotation is valuable but costly for large-scale evaluation. We thus scale up the dataset using a semi-automated procedure, transforming knowledge triples in Wikidata (2022-10-03) to natural language questions with human-authored templates, resulting in 1.3M examples over 11K visual entities covering 2.7K entity types (see Table~\ref{tab:datasets}).

\custompara{QA Generation.} 
We convert knowledge triples ({\small \texttt{subj}, \texttt{relation}, \texttt{obj}}{}) in Wikidata to natural language question-answer pairs for a selected list of 300 relations. For each relation, annotators write one or two question templates, which contain a placeholder for a hypernym of the visual entity (e.g., car) and a placeholder for unit measurements (e.g., inches) in numerical questions to avoid ambiguity. 
Finally, we construct the IQA triples by pairing images of a visual entity with corresponding QA pairs.\footnote{Based on manual inspection of 500 examples, we found this process rarely produces incorrect examples ($\le$1.2\%).}

\custompara{QA Pair Filtering and Subsampling.}
To ensure the questions are diverse and the answers can be referenced from Wikipedia, we filter out QA pairs when answers from Wikidata cannot be found in the Wikipedia article and subsample questions to balance the distribution of entities and relations.

\subsection{Evaluation of \textbf{\ourdataset}}
\label{sec:split}

\custompara{Dataset Split.}
We design the evaluation split to prevent overfitting to the training set and focus on evaluating the generalization ability of the pre-trained models. This includes the ability to answer questions of new entities and questions not seen during training.
Particularly, we define two evaluation splits: (1) \textsc{Unseen Entity}, where a portion of entities are held out during training and only included in the evaluation; (2) \textsc{Unseen Question}, where we hold out a portion of the QA pairs of seen entities for evaluation. 

\custompara{Evaluation Metric.}
\label{sec:metric}
Three types of questions are evaluated differently: VQA accuracy~\cite{balanced_vqa_v2} for \textsc{String} and \textsc{Time}; {\em Relaxed Accuracy}~\cite{methani2020plotqa} for \textsc{Numerical}.
We applied different relaxing strategies for each question type, and averaged the accuracy for each question.
Finally, we calculate the accuracy for each data split (\textsc{Unseen Question} and \textsc{Unseen Entity}), and take the harmonic mean of them as the overall accuracy (see Appendix).

\section{Protocols and Models for {\infoseek}}
\label{sec:method}
Motivated by previous research on text-based question benchmarks~\cite{joshi-etal-2017-triviaqa,roberts2020much}, we introduce two evaluation protocols, \ie, \emph{No KB} and \emph{With KB}, to evaluate models with different information accessible from {\sc InfoSeek}.
Table~\ref{tab:eval_protocols} and Figure~\ref{fig:models} have provided a comparison for the two setups.
This key design choice is made to encourage models from different families to be compared with a clear notion of what information was accessed. We note that the \emph{No KB} protocol is more challenging than the \emph{With KB} protocol.

\custompara{The \nokb protocol.} Models are tasked to directly predict the answer by examining the image and question, similar to traditional VQA systems. 
This requires the model to store world knowledge in its parameters for effective question answering. The research question focuses on how much knowledge can an end-to-end model memorize in its parameters during pre-training, and how well can it utilize this knowledge after fine-tuning? 
We use the standard VQA formatted data, \ie, \{Image ($\mathsf{I}$), Question($\mathsf{Q}$), Answer($\mathsf{A}$)\} triplets for training / validation, and \{$\mathsf{I}, \mathsf{Q}$\} for testing.

\custompara{The \withkb protocol.} The goal is to analyze headroom for improvement when a viable reasoning chain is explicitly provided.
Therefore, this protocol encourages an extra step of visual entity recognition, grounding the task on a knowledge base.
The VQA task is transformed into a two-step pipeline, \ie, (1) visual entity recognition; and (2) language QA with entity information. 
We provide training signals to first recognize the queried visual entity and then leverage the information to query a large language model for answers, or identify relevant Wikipedia articles for extracting the answer. 
Specifically, we provide a 100K Wikipedia KB (articles and infobox images) that includes visual entities from \ourdataset and top frequent entities from Wikipedia.
During training and validation, \emph{With KB} protocol provides entity labels for each queried visual entity. 
During testing, the model is evaluated based on the \{$\mathsf{I}, \mathsf{Q}$\} pairs only.

\subsection{Models without KB Information}
\label{subsec:e2e_models}
\custompara{Random \& Prior.} Random answers sampled from the training set; The majority answer based on the question prior, which is calculated using the training set questions grouped by question 4-gram.

\custompara{\qonly~Model.}
To validate the importance of visual content in {\sc InfoSeek}, we build a question-only baseline with PaLM (540B)~\citep{chowdhery2022palm}, using text question as the only input and with 5-shot in-context-learning.

\custompara{BLIP2~\&~InstructBLIP.}
We utilize two pre-trained vision-language models, \ie, BLIP2~\citep{li2023blip} and InstructBLIP~\citep{dai2023instructblip}.
Both models share the same architecture, which trains a Q-former Transformer that connects a frozen vision encoder (\snlp{ViT-g/14}) to a frozen instruction-tuned language model (\snlp{Flan-T5$_{\text{XXL}}$}~\cite{chung2022scaling}) to output text based on an input image and text. Particularly, InstructBLIP fine-tunes the BLIP2 model on 26 vision-language datasets (\eg, \snlp{VQAv2, OKVQA}) with a text instruction prefix, and claimed to show improved zero-shot performance on unseen vision-language tasks.
Following \citet{li2023blip}, we fine-tune the Q-former of both models using the \infoseek$_{\text{Wikidata}}$, for improved performance.

\custompara{PaLI-17B \& PaLI-X.}
We experiment with two extra pre-trained vision-language models from the PaLI~\citep{chen2022pali,chen2023pali} family given its SOTA performance.
Particularly, we use PaLI-17B (ViT-e + mT5$_{\text{XXL}}$~\citep{xue2020mt5}) and PaLI-X (ViT-22B~\citep{dehghani2023scaling} + UL2-33B~\citep{tay2022unifying}), which are pre-trained on WebLI~\citep{chen2022pali} with 1 billion image-text pairs.
Both models, which use non instruction-tuned language models, exhibit minimal zero-shot performance on \infoseek.
Consequently, we fine-tune both models on the \infoseek$_{\text{Wikidata}}$ to improve their performance.

\begin{table*}[!tbh]
\small
\tabcolsep 2.5pt
\begin{tabular}{lcccccccc|cc}
\toprule
\multirow{3}{*}{\textbf{Model}}  &
\multirow{3}{*}{\textbf{LLM}} &\multirow{3}{*}{\textbf{\# Params}} & \multicolumn{3}{c}{$\textbf{\ourdataset}_{\text{Wikidata}}$} & \multicolumn{3}{c}{$\textbf{\ourdataset}_{\text{Human}}$} & \textbf{OK-VQA} & \textbf{VQAv2}\\
\cmidrule(lr){4-6}\cmidrule(lr){7-9}\cmidrule(lr){10-10}\cmidrule(lr){11-11}
 & & & \small \textsc{Unseen}  & \small \textsc{Unseen} & \multirow{2}{*}{\small Overall} & \small \textsc{Unseen} & \small \textsc{Unseen}  & \multirow{2}{*}{\small Overall} & \multirow{2}{*}{\small Accuracy} & \multirow{2}{*}{\small Accuracy}\\
 & & & \small \textsc{Question} &\small  \textsc{Entity} & & \small \textsc{Question} &  \small \textsc{Entity} &  & &\\
\midrule
Random & - & - & 0.1 & 0.1 & 0.1 & 0.2 & 0.1 & 0.1 & -\pz & -\pz\\
Prior & - & - & 3.9	&2.7&3.2 &0.3 & 0.3 & 0.3 & -\pz & 32.1 $^\dagger$ \\
\qonly & PaLM & 540B & 5.1 & 3.7 & 4.3 & 4.8 & 6.6 & 5.6 & 23.8\pz & 43.0\pz \\ \midrule
BLIP2 & Flan-T5$_{\textsc{xxl}}$ & \pz12B & 14.5& 13.3& 13.9& 10.0& 8.9& 9.4& 54.7\pz& 82.3\pz \\
InstructBLIP & Flan-T5$_{\textsc{xxl}}$ & \pz12B & 14.3& 13.2& 13.7& 10.6& 9.3& 9.9& 55.5\pz& -\pz \\
\palift-17B & mT5$_{\textsc{xxl}}$ & \pz17B & 20.7 &	16.0	& 18.1 & 13.3 & 5.9 & 8.2 & 64.8\pz & 84.6\pz \\
PaLI-X & UL2$_{32\textsc{B}}$ & \pz55B & 23.5 & 20.8 & 22.1& 12.9& 9.3& 10.8& 66.1\pz & 86.1\pz \\
\bottomrule
\end{tabular}

{\scriptsize $\dagger$: Numbers adopted from~\citeauthor{agrawal2018don}}

\caption{
\textbf{Results of No-KB models fine-tuned on \ourdataset.} Baselines including Random, Prior (majority answer with 4-gram question prior), and a question-only model using PaLM (Q-only) with 5-shot prompting. 
VQA accuracy of models on OK-VQA~\cite{marino2019ok} and VQAv2~\cite{balanced_vqa_v2} are for comparison.
}
\label{tab:withoutkb}
\end{table*}

\subsection{Models with KB Information}
\label{subsec:pipeline_models}

In this protocol, we explicitly model the path to answer info-seeking questions with two decoupled sub-tasks: (1) recognizing the visual entity grounded to the KB and (2) textual reasoning to answer the question. A hidden benefit of such pipeline systems is improved interpretability, because it is easier to locate the source of errors by diagnosing each sub-task component.

\custompara{Sub-task \#1: Visual Entity Recognition.}
We follow the entity recognition task defined in OVEN~\cite{hu2023opendomain}, and use an image and a text query (e.g., ``\nlp{What is this building?}") as model inputs, and predict entities among 100K multi-modal Wikipedia entries.
Particularly, we employ the pre-trained CLIP~\cite{radford2021clip} model (\snlp{ViT-L/14}), as our visual entity recognition model, because of its strong generalization capability.
Specifically, we follow the CLIP2CLIP model described in~\citeauthor{hu2023opendomain}, to fine-tune CLIP to encode multi-modal representations (\snlp{image, question}) from our dataset as query, and (\snlp{Wikipedia image, Wikipedia title}) from the KB as candidates.
We then retrieve the top \textit{k}=5 most similar entities based on weighted cosine similarity scores computed between the query and candidates. 

\custompara{Sub-task \#2: Language QA with LLM or KB Reader.}
Through visual entity recognition, we can now represent the queried visual information as its textual description.
This enables us to investigate the language reasoning component independently to understand how much improvement a strong LLM or a KB reader can bring.

\begin{itemize}[leftmargin=*,topsep=0pt,itemsep=0pt]
    \item \textbf{PaLM: Large Language Model.}
    We use PaLM (540B) to investigate the amount of knowledge that can be memorized in the model's parameters from pre-training on text corpora.
    Given a question and the queried entity name (from entity recognition), we prompt PaLM to predict the answer using 5-shot in-context examples with the prompt format: \nlp{``question: This is \{entity\} \{question\} answer:''}.
    \item \textbf{Fusion-in Decoder (FiD): KB Reader.}
    We experiment with a SOTA retrieval-augmented model, which reads information from a KB, to understand the value of Wikipedia articles in the KB.
    Specifically, the FiD~\cite{izacard2020fid} model is employed, which takes \textit{N}=100 retrieved articles as input and generates an answer. 
    The model is pre-trained with a T5$_\text{Large}$~\cite{2020t5} backbone (660M) on Natural Questions~\cite{kwiatkowski-etal-2019-natural} and fine-tuned on {\sc InfoSeek}. 
    During inference, we retrieve the first 20 passages from Wikipedia for $k$=5 visual entities (from entity recognition) and feed 100 passages to FiD to generate the answer. 
\end{itemize}

\section{Experiments}
\label{sec:experiment}
\subsection{Results for No-KB Models}
\label{subsec:results_withnokb}
\custompara{Main results.} 
Table~\ref{tab:withoutkb} presents the results of end-to-end models on \infoseek. 
The best pre-trained model in this setting is PaLI-X, although the absolute number on the model's overall performance remains low.
This is partially due to the fact that \infoseek questions often require identifying entities and retrieving specific information relevant to the question, making it a challenging task for end-to-end models.
As PaLI-X is pre-trained on a large corpus with more model parameters, it demonstrates better generalization ability on the \textsc{unseen entity} split compared to PaLI-17B. 
Meanwhile, there remains a noticeable gap in performance on the \textsc{unseen question} and \textsc{unseen entity} splits, indicating that models struggle with generalization to new visual entities from the training set.
We also present models' results on OK-VQA~\cite{marino2019ok} and VQAv2~\cite{balanced_vqa_v2} for comparison and observe a drastic performance gap, emphasizing the difficulty of visual info-seeking questions.

\custompara{Fine-tuning elicits knowledge from the model.} 
To demonstrate the value of \ourdataset training data, we report the zero-shot performance of models in Figure~\ref{fig:zeroshot}. 
Specifically, we find that without fine-tuning, both PaLI models produce a negligible overall performance, which is significantly worse than the fine-tuned counterpart.
This provides evidence to support the hypothesis that fine-tuning has helped elicit knowledge from the pre-trained PaLI models. 
On the other hand, BLIP2 and InstructBLIP show compelling zero-shot performance on \ourdataset as they adopt a frozen instruction fine-tuned LLM (\ie,  \snlp{Flan-T5}) and InstructBLIP is further instruction-tuned on a collection of VQA benchmarks.
The performance of BLIP2 models is further improved after fine-tuning on \ourdataset with a small number of steps, showing strong generalization results to the Human split. In Figure~\ref{fig:blip2_generalization}, we present examples of BLIP2 predicting the ``country location'' of an unseen entity (\ie Amberd) and show the accuracy was improved from 18\% to 92\% after fine-tuning, despite not seeing this entity in the training set.
Finally, we conducted a real-world evaluation on out-of-domain images unavailable from the Internet (not from any models' pre-training data).
Particularly, we evaluate fine-tuned PaLI with 90 questions on 30 images captured by the authors, on visual entities outside of the \ourdataset training corpus. 
As a result, PaLI-17B and PaLI-X answered 22.2\% and 38.9\% of questions correctly.
Figure~\ref{fig:human_demo} presents examples of PaLI and BLIP2 predictions on two out-of-domain entities (artwork and fashion product).

\begin{figure}[t]
    \centering
    \begin{tikzpicture}
    \begin{axis}[
        width=1\columnwidth,
        height=0.6\columnwidth,
        ybar,
        ymin=0,
        ymax=25,
        xmin=-0.5,
        xmax=7.5,
        bar width=2mm,
		xlabel near ticks,
		ylabel near ticks,
        xtick=data,
        font=\small,
        xticklabels={PaLI-17B, PaLI-X, BLIP2, InstructBLIP, PaLI-17B, PaLI-X, BLIP2, InstructBLIP},
        x tick label style={rotate=45,anchor=east},
        ylabel={Accuracy \%},
        axis lines={left},
        axis line style={-{}},
        legend columns=1,
        legend style={at={(1,0.85)}, anchor=north east}
    ]
    \draw [dashed] (axis cs:3.5,\pgfkeysvalueof{/pgfplots/ymin}) -- (axis cs:3.5,\pgfkeysvalueof{/pgfplots/ymax});
    \addplot[fill=red!30] coordinates {
        (0, 2.0)
        (1, 2.0)
        (2, 11.3)
        (3, 7.4)
        (4, 1.7)
        (5, 1.6)
        (6, 8.6)
        (7, 7.2)
    };
    \addplot[fill=citecolor!50] coordinates {
        (0, 18.1)
        (1, 22.1)
        (2, 13.9)
        (3, 13.7)
        (4, 8.2)
        (5, 10.8)
        (6, 9.4)
        (7, 9.9)
    };
    \legend{Zero-shot, Fine-tuned}
    \node at (axis cs:1.5,24) {\ourdataset$_\text{Wikidata}$}; 
    \node at (axis cs:5.5,24) {\ourdataset$_\text{Human}$}; 
    \end{axis}
    \end{tikzpicture}
    \vspace{-2.5mm}
    \caption{\textbf{Zero-shot \& fine-tuned performances on {\sc InfoSeek}}. Fine-tuning on \infoseek~elicits knowledge from PaLI models to answer fine-grained visual info-seeking questions.  
    }
    \vspace{-2mm}
    \label{fig:zeroshot}
\end{figure}
\definecolor{myred}{RGB}{204,41,54}
\definecolor{mygreen}{RGB}{0,128,0}
\begin{figure}
    \scriptsize
  \centering
  \begin{minipage}[t]{0.22\textwidth}
    \centering
    \begin{tikzpicture}
        \node [inner sep=0pt] at (0,0) {\includegraphics[width=\linewidth ]{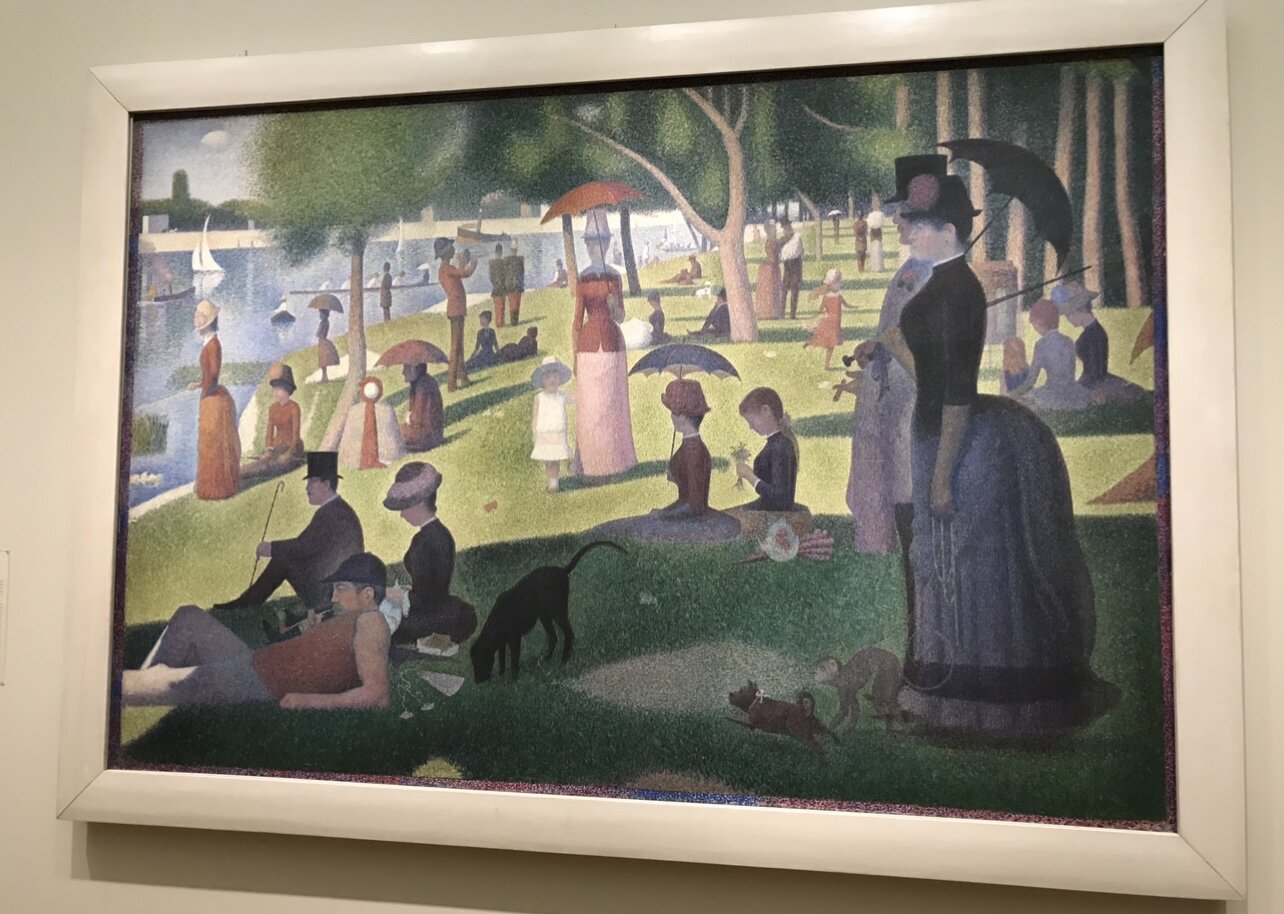}};
        \draw [white, rounded corners=\ClipSep, line width=\ClipSep] 
            (current bounding box.north west) -- 
            (current bounding box.north east) --
            (current bounding box.south east) --
            (current bounding box.south west) -- cycle
            ;
    \end{tikzpicture}
    \begin{itemize}[nosep, align=left, leftmargin=*,label={}]
      \item \textsf{\textbf{Q}: what year was this painting created?}
      \item \textsf{\texttt{PaLI-17B}: 1884} {\color{mygreen}\cmark}
      \item \textsf{\texttt{PaLI-X}: 1884} {\color{mygreen}\cmark}
      \item \textsf{\texttt{BLIP2}: 1887}
      {\color{myred}\xmark}
    \end{itemize}
  \end{minipage}
  \hfill
  \begin{minipage}[t]{0.22\textwidth}
    \centering
    \begin{tikzpicture}
        \node [inner sep=0pt] at (0,0) {\includegraphics[width=\linewidth ]{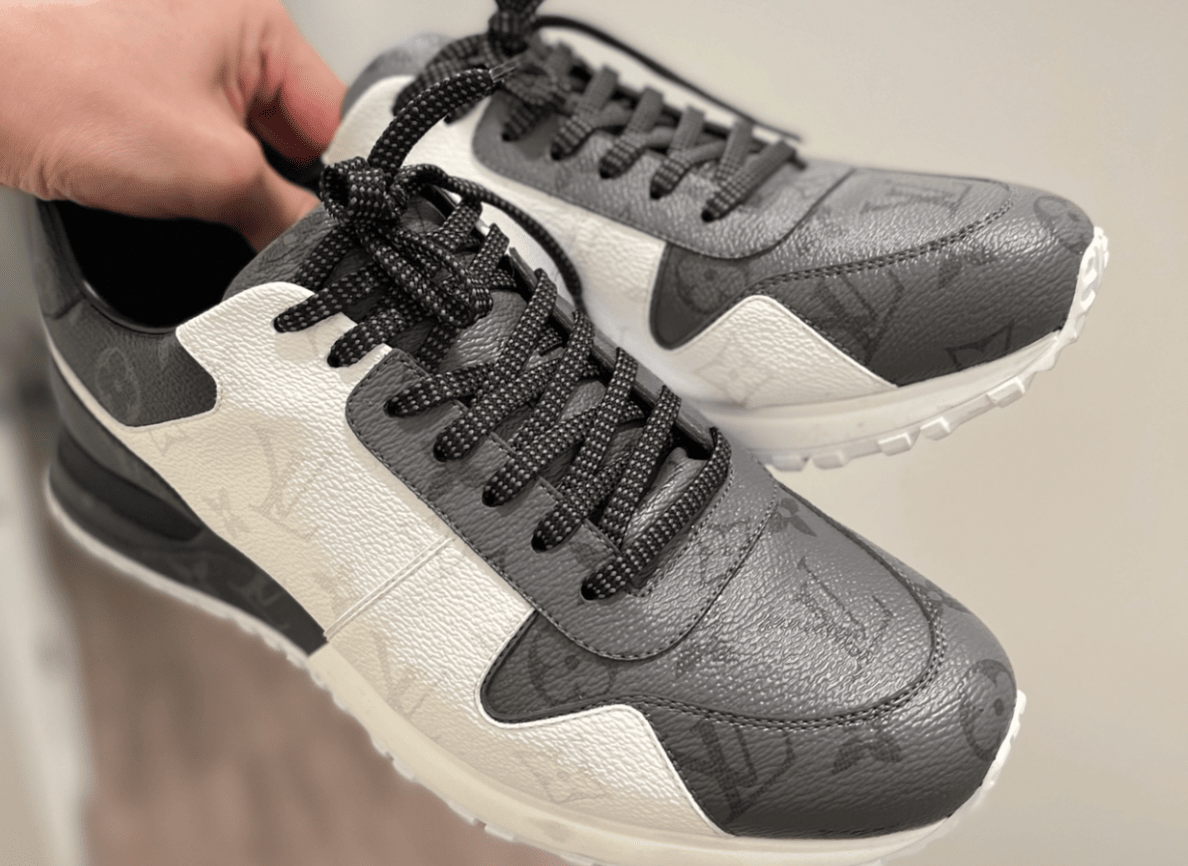}};
        \draw [white, rounded corners=\ClipSep, line width=\ClipSep] 
            (current bounding box.north west) -- 
            (current bounding box.north east) --
            (current bounding box.south east) --
            (current bounding box.south west) -- cycle
            ;
    \end{tikzpicture}

    \begin{itemize}[nosep, align=left, leftmargin=*,label={}]
      \item \textsf{\textbf{Q}}: \textsf{which year was this brand established?}
      \item \textsf{\texttt{PaLI-17B}: 1915} {\color{myred}\xmark}
      \item \textsf{\texttt{PaLI-X}: 1854} {\color{mygreen}\cmark}
      \item \textsf{\texttt{BLIP2}: 1854}
      {\color{mygreen}\cmark}
    \end{itemize}
  \end{minipage}
  \caption{\textbf{Predictions on out-of-domain visual entities} (art \& fashion) collected from real-world images by authors, using \ourdataset fine-tuned models.}
  \label{fig:human_demo}
\end{figure}

\custompara{Why does instruction-tuned BLIP2 obtain worse zero-shot \ourdataset results?}
One surprising finding from Figure~\ref{fig:zeroshot} caught our attention and reveals an important criterion to be considered for future model development.
We found InstructBLIP$_{\texttt{0-shot}}$ performs significantly worse than its initial checkpoint, BLIP2 (\snlp{7.4 vs 11.3 on \infoseek$_{\text{Wikidata}}$}), which contradicts the superior zero-shot performances of InstructBLIP in \citet{dai2023instructblip}.
We conduct manual analysis and detect a common error made by InstructBLIP is its preference for generating coarse-grained predictions compared to BLIP2 (\eg, \snlp{architect vs a person's name}).
This leads to a performance drop on \infoseek, which emphasizes fine-grained answers (see Figure~\ref{fig:instruct_error}). 
We hypothesize that this can be attributed to the instruction tuning datasets used for InstructBLIP (\eg, \snlp{VQAv2 and OK-VQA}), which share a less fine-grained answer distribution.
Fortunately, fine-tuning on \infoseek$_\text{Wikidata}$ helps close the gap.

\definecolor{myred}{RGB}{204,41,54}
\begin{figure}
  \centering
  \scriptsize
  \begin{minipage}[t]{0.22\textwidth}
    \centering
    \begin{tikzpicture}
        \node [inner sep=0pt] at (0,0) {\includegraphics[width=\linewidth]{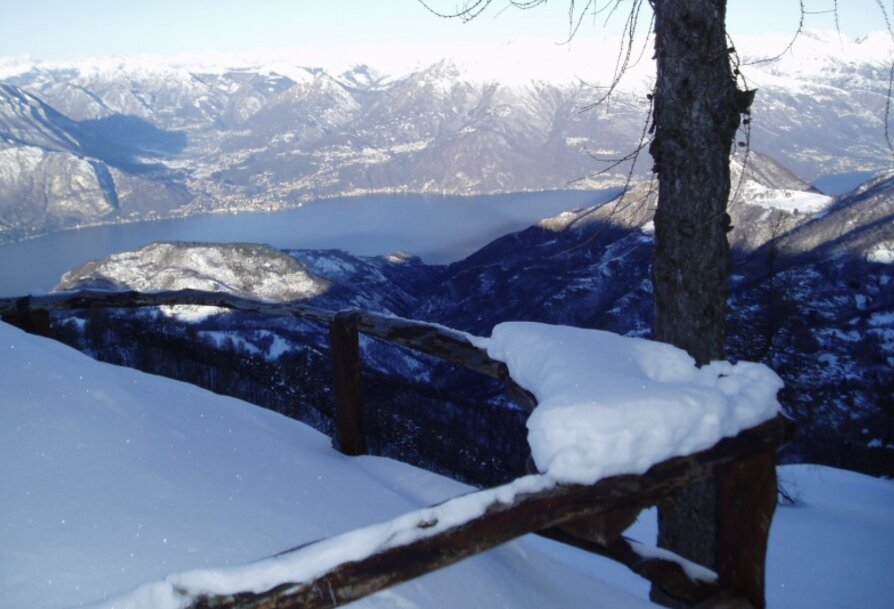}};
        \draw [white, rounded corners=\ClipSep, line width=\ClipSep] 
            (current bounding box.north west) -- 
            (current bounding box.north east) --
            (current bounding box.south east) --
            (current bounding box.south west) -- cycle
            ;
    \end{tikzpicture}

    \begin{itemize}[nosep, align=left, leftmargin=*,label={}]
      \item \textsf{\textbf{Q}:  Which body of water is this mountain located in or next to?}
      \item \textsf{\textbf{A}:  Lake Como}
      \item \textsf{\texttt{BLIP2}$_\texttt{(0-shot)}$: lake como}
      \item \textsf{\texttt{InstructBLIP}$_\texttt{(0-shot)}$: \textcolor{myred}{lake}}
    \end{itemize}
  \end{minipage}
  \hfill
  \begin{minipage}[t]{0.25\textwidth}
    \centering
    \begin{tikzpicture}
        \node [inner sep=0pt] at (0,0) {\includegraphics[width=0.9\linewidth ]{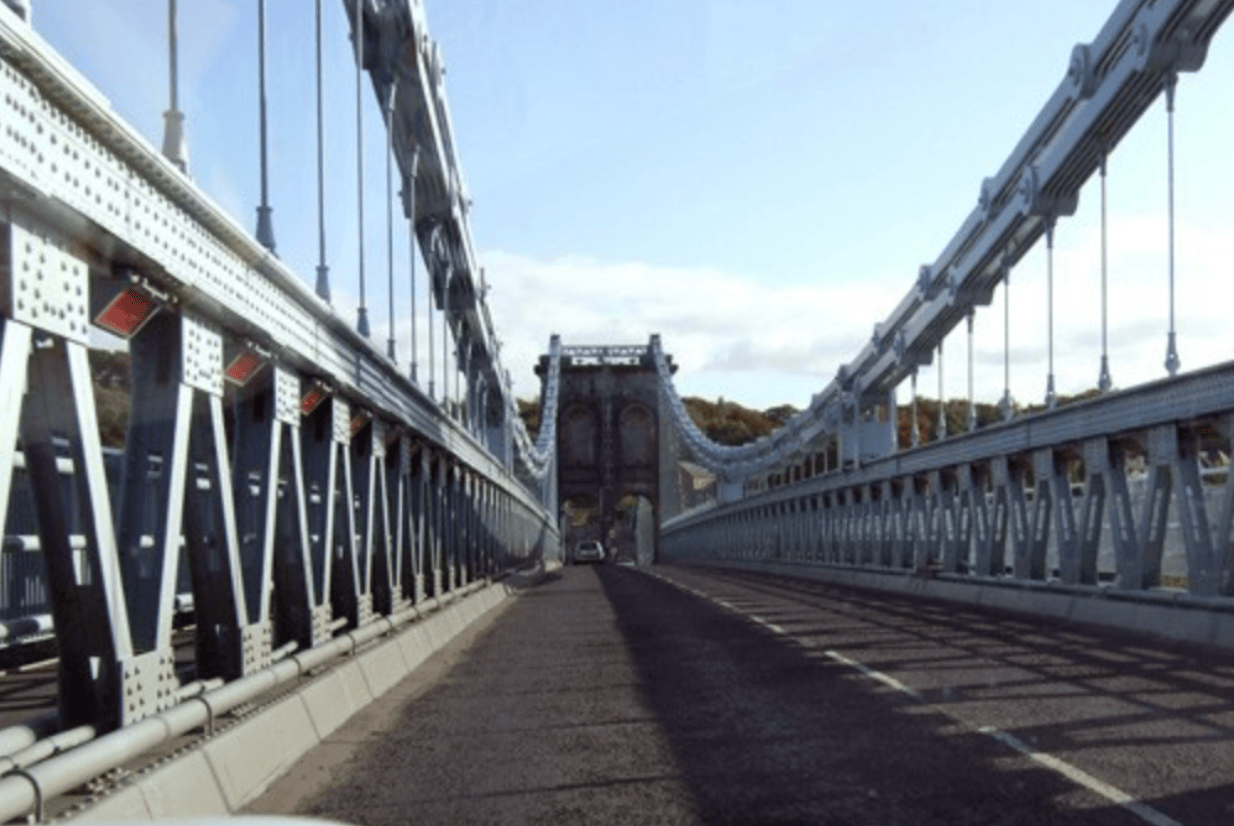}};
        \draw [white, rounded corners=\ClipSep, line width=\ClipSep] 
            (current bounding box.north west) -- 
            (current bounding box.north east) --
            (current bounding box.south east) --
            (current bounding box.south west) -- cycle
            ;
    \end{tikzpicture}
    
    \begin{itemize}[nosep, align=left, leftmargin=*,label={}]
      \item \textsf{\textbf{Q}: Who designed this bridge?}
      \item
      \item \textsf{\textbf{A}: Thomas Telford}
      \item \textsf{\texttt{BLIP2}$_\texttt{(0-shot)}$: john nash}
      \item \textsf{\texttt{InstructBLIP}$_\texttt{(0-shot)}$: \textcolor{myred}{architect}}
    \end{itemize}
  \end{minipage}
  \vspace{-2mm}
  \caption{\textbf{InstructBLIP}$_\texttt{(0-shot)}$ \textbf{makes less fine-grained predictions} compared to its initial model (BLIP2), after instruction-tuned on prior VQA datasets.}
  \label{fig:instruct_error}
  \vspace{-4mm}
\end{figure}
\subsection{Results for With-KB Models}
\label{subsec:results_withkb}

\begin{table}[!htbp]
\centering
\footnotesize
\tabcolsep 6pt
\begin{tabular}{lcc|c}
\toprule
\multirow{2}{*}{Model} & \ourdataset & \ourdataset & \textsc{Entity}\\
& \tiny Wikidata & \tiny Human & \tiny Accuracy\\
\midrule
Best No-KB & 22.1 & 10.8 & - \\
\midrule
\textbf{With-KB Setting}\\
CLIP $\to$ PaLM &  20.1 &  15.2 & \multirow{2}{*}{22.2}  \\
CLIP $\to$ FID &  19.3 & 18.2 &  \\
\rowcolor{lightgray} Oracle $\to$ FID &  52.0 & 45.6 & 100 \\
\bottomrule
\end{tabular}
\caption{\textbf{Results of With-KB setting.}
CLIP $\to$ PaLM/FID: a two-stage pipeline system (visual entity recognition $\to$ text QA).
PaLM: 5-shot prompting.
FID: Fusion-in Decoder to read from KB using T5$_{\text{large}}$.
Oracle: an artificial upper-bound using oracle entities.}
\label{tab:oven}
\end{table}

\custompara{Models with KB access perform better.}
Table~\ref{tab:oven} presents the results for pipeline models with access to knowledge base (KB) information, along with the best results from the No-KB setting for reference. Notably, the pipeline models outperform the best No-KB models on the challenging \infoseek$_{\text{Human}}$ split significantly. 
This highlights the pipeline systems' ability to answer visual info-seeking questions by effectively utilizing visual recognition and language reasoning, specifically using the names of visual entities to convey information across modalities.
When comparing the two Language QA models, we observe that the FiD model, which reads the Wikipedia article, achieves the highest generalization performance on \infoseek$_{\text{Human}}$ by a significant margin. This suggests that access to relevant text content plays a crucial role in answering visual info-seeking questions.

\custompara{Large headroom for improvement.} 
Table~\ref{tab:oven} demonstrates an artificial upper-bound (\snlp{Oracle $\to$ FiD}) on \ourdataset, indicating substantial room for performance improvement if an oracle entity recognition model were available. By simulating the visual entity recognition's accuracy improvement (\snlp{from 22\% using CLIP to 100\%}), the \ourdataset accuracy can be improved from $\sim$20\% to $\sim$50\%, within the same FiD model.

\custompara{Analysis on each question type.}
\begin{table}[!tb]
\centering
\small
\tabcolsep 5pt
\begin{tabular}{lccc}
\toprule
\multirow{2}{*}{\textbf{Model}}&  \small \textsc{Time} & \small \textsc{Numerical} & \small \textsc{String} \\
 & \scriptsize (Acc.) & \scriptsize (Relaxed Acc.) & \scriptsize (Acc.)\\
\midrule
\textbf{\nokb Setting}\\
Prior & \pz\pz0	& \pz4.4	& 5.0 \\
\qonly & \pz\pz0	&11.4&	4.0 \\
InstructBLIP & \pz7.9 & \pz7.5 & 17.8 \\
BLIP2 & \pz6.9 & \pz5.8 & 18.5 \\
\palift-17B & \pz3.8	& 18.4	& 27.4\\
PaLI-X & \pz7.7	& 16.1	&30.0\\
\midrule
\textbf{\withkb Setting}\\
CLIP $\to$ PaLM & 12.5	& 27.7	&21.7\\
CLIP $\to$ FiD & 12.3	& 23.4	& 23.9\\
\bottomrule
\end{tabular}
\vspace{-2mm}
\caption{\textbf{Results w.r.t. each question types} on the ${\ourdataset}_{\text{Wikidata}}$ val set of unseen question split, showing a big headroom for improvements on \textsc{Time} and \textsc{Numerical} for all end-to-end models.
}
\label{tab:question_type}
\end{table}
Table~\ref{tab:question_type} shows a breakdown of results under different question types, evaluated on \ourdataset$_{\text{Wikidata}}$. 
Comparing \textbf{No KB} and \textbf{With KB} models, we found that end-to-end models such as PaLI, have a short barrel on fine-grained knowledge-intensive questions (\ie, \textsc{Time} and \textsc{Numerical}).
It can perform well on other questions, which are more about querying attributes or resolving relations between entities (see Figure~\ref{fig:question_relation}). 
Comparing \textbf{With KB} models, PaLM and FiD perform on par with each other on this automated evaluation data. However, when evaluated on the natural info-seeking human queries, FiD has a better generalization, outperforming PaLM on \textsc{Time} (21.5 vs 14.6) and \textsc{Numerical} (25.6 vs 21.3) questions from \infoseek$_{\text{Human}}$ significantly.
One possible reason is that natural info-seeking questions written by people focus more on very fine-grained information, which is rare and hard to memorize for PaLM. 
In contrast, FiD can leverage Wikipedia articles to predict answers. 
Finally, we analyze the performance of different models according to the visual entity popularity and found unique advantages of end-to-end models (see Appendix).

\begin{figure}[t]
    \centering
    \begin{tikzpicture}
    \begin{axis}[
        width=0.85\columnwidth,
        height=0.5\columnwidth,
        xbar,
        xmin=0,
        xmax=90,
        ymin=-0.5,
        ymax=3.5,
        bar width=1.5mm,
		ylabel near ticks,
		xlabel near ticks,
        ytick=data,
        font=\small,
        yticklabels={
        \raisebox{-8mm}{\parbox{2.5cm}{\textsf{\scriptsize What is the mountain range..?}}},
        \textsf{\scriptsize Which continent..?}, 
        \textsf{\scriptsize Which country..?}, 
        \textsf{\scriptsize Which brand..?}},
        y tick label style={rotate=0,anchor=east,text width=2.5cm,align=left},
        ylabel={},
        xlabel={Accuracy \%},
        axis lines={left},
        axis line style={-{}},
        legend style={at={(1.025,0.25)}, font=\footnotesize, anchor=east}
    ]
    \addplot[fill=red!30] coordinates {
        (15.3, 0)
        (23.0, 1)
        (38.4, 2)
        (65.7, 3)
    };
    \addplot[fill=citecolor!50] coordinates {
        (34.7, 0)
        (56.1, 1)
        (66.3, 2)
        (87.1, 3)
    };
        \legend{With-KB, No-KB}
    \end{axis}
    \end{tikzpicture}
    \caption{No-KB (\snlp{PaLI-17B}) outperforms With-KB (\snlp{CLIP$\to$FiD}) models on questions that query less fine-grained attributes.}
    \label{fig:question_relation}
\end{figure}

\custompara{Performance on Head vs. Tail entities.}
Although pipeline models with KB access are overall stronger, surprisingly, we observe that end-to-end models have a unique advantage for info-seeking VQA, particularly on the tail entities. 
Figure~\ref{fig:entity_popular} presents a comparison of models, with group-wise performances on Wikipedia entities that are least popular (less monthly page views) to most popular (more monthly page views). The histogram is generated based on the average monthly Wikipedia pageviews in 2022, following~\cite{mallen2022not}.
Surprisingly, the results show that PaLI-17B outperforms the pipeline systems by a large margin on the tail entities, particularly for questions related to geographical information. We show some qualitative examples in Figure~\ref{fig:error_analysis}, for entities from baskets of different monthly page views. 
This suggests that there are many different routes to answer visual info-seeking questions and that pipeline systems that rely on an explicit decomposition of the VQA task may be redundant and susceptible to error propagation from the entity linking stage.
Whereas for end-to-end models such as PaLI, it is flexible to decide which route of reasoning is more appropriate to answer a given question. For example, one can answer geographical questions without knowing the identity of the visual entity, if other relevant visual clues are presented.
Meanwhile, on the more popular head visual entities, a clear trend emerged showing that pipeline systems outperform end-to-end PaLI by a big margin.
\definecolor{pali}{HTML}{feebe2}
\definecolor{palm}{HTML}{f768a1}
\definecolor{fid}{HTML}{c51b8a}

\begin{figure}[t]
    \centering
    \begin{tikzpicture}
    \begin{axis}[
        width=1\columnwidth,
        height=0.6\columnwidth,
        ybar,
        ymin=0,
        ymax=40,
        xmin=-0.5,
        xmax=9.5,
        bar width=1mm,
		xlabel near ticks,
		ylabel near ticks,
		ylabel style={yshift=-5pt},
        xtick=data,
        font=\small,
        xticklabels={100, 500, 1K, 2.5K, 5K, 10K, 25K, 50K, 100K, 1M},
        xlabel={Average Monthly View of Entity},
        ylabel={Accuracy \%},
        axis lines={left},
        axis line style={-{}},
        legend columns=3,
        legend style={at={(1.02,1)}, font=\small},
    ]
    \addplot[fill=pali] coordinates {
        (0, 21.9)
        (1, 19.5)
        (2, 21.5)
        (3, 15.3)
        (4, 20.4)
        (5, 19.4)
        (6, 23.6)
        (7, 17.9)
        (8, 16.3)
        (9, 18.7)
    };
    \addplot[fill=palm!80] coordinates {
        (0, 11.4)
        (1, 10)
        (2, 10.7)
        (3, 15.1)
        (4, 20.3)
        (5, 25.8)
        (6, 25.9)
        (7, 27.4)
        (8, 21.2)
        (9, 20.9)
    };
    \addplot[fill=fid!80] coordinates {
        (0, 10.1)
        (1, 17.2)
        (2, 17.1)
        (3, 17.6)
        (4, 23.3)
        (5, 23.4)
        (6, 24.3)
        (7, 22.4)
        (8, 19.4)
        (9, 19.9)
    };
    \legend{PaLI-17B, CLIP$\rightarrow$PaLM, CLIP$\rightarrow$FiD}
    \end{axis}
    \end{tikzpicture}
    \caption{\textbf{\ourdataset results w.r.t. visual entities of different popularity.} End-to-end model outperforms pipeline systems on tail entities (low monthly pageviews) but overturned on more popular entities (high monthly pageviews). 
    }
    \label{fig:entity_popular}
\end{figure}

\section{Related Work}
\label{sec:related}
\custompara{Pre-trained Vision Language Models.}
There has been significant growth in the development of vision-language models pre-trained on large-scale image-text datasets~\cite{lu2022unified,bao2021beit,wang2022unifying,zhou2020unified,radford2021clip}. 
One line of research aims to augment a pre-trained language model with visual modality by learning a mapping from an external visual encoder to the frozen large language model~\cite{alayrac2022flamingo,li2023blip,koh2023grounding}, to fully leverage textual knowledge from the language model~\citep{xu2022multiinstruct,dai2023instructblip,liu2023visual,zhu2023minigpt,ye2023mplug}.

\custompara{Knowledge-based VQA Models.}
Various approaches have been proposed to address knowledge-based VQA tasks~\cite{marino2019ok} by incorporating external knowledge into vision-language models. One approach is to retrieve information from an external KB~\cite{marino2021krisp, hu2022reveal, wu2022entity} and employ a model~\cite{izacard2020fid} to perform language QA~\cite{gui2021kat, lin2022revive}. Other approaches transform the image into a text caption and use an LLM~\cite{brown2020language, chowdhery2022palm} to answer questions~\cite{yang2022pica, hu2022promptcap}.
We utilize both approaches to study the ceiling for improvement on \infoseek with the OVEN model~\cite{hu2023opendomain}.

Another concurrent work~\citep{mensink2023encyclopedic} investigates similar challenges but emphasizes scalability and relies on model-generated annotations, as opposed to our human-annotated info-seeking queries.

\section{Conclusion}
\label{sec:conclusion}
We introduced \infoseek, a large-scale VQA dataset that focuses on answering visual information seeking questions. 
With {\sc InfoSeek}, we found that current state-of-the-art pre-trained visual-language models struggle to answer visual info-seeking questions requiring fine-grained knowledge, such as questions about time and numerical information of a visual entity.
Our analysis using pipeline systems, which ground visual entities to an external knowledge base, suggests that incorporating fine-grained knowledge into the pre-training process holds significant potential to improve end-to-end pre-training models. 

\section{Limitation}
\label{sec:limitation}
\ourdataset is limited to English language and future research could expand it to a multilingual setting, leveraging articles in Wikipedia supported in other languages.
While the primary focus of this work is on knowledge derived from Wikipedia, future investigations could explore extensions to other domains, such as medical information, and artwork, and incorporate emerging updates in Wikipedia~\cite{iv-etal-2022-fruit}.

\section*{Acknowledgement}
We thank Jialin Wu, Luowei Zhou for reviewing an early version of this paper. We thank Xi Chen for providing different variants of PaLI pre-trained checkpoints. We also thank Radu Soricut, Anelia Angelova, Fei Sha, Andre Araujo, Vittorio Ferrari, Wei Xu, Kartik Goyal for valuable discussions and feedback on the project. Yang Chen is partially funded by the NSF (IIS-2052498).

\clearpage
\newpage

\bibliography{custom}
\bibliographystyle{acl_natbib}

\clearpage
\newpage

\appendix
\begin{figure*}[ht!]
    \centering
    \includegraphics[width=\textwidth]{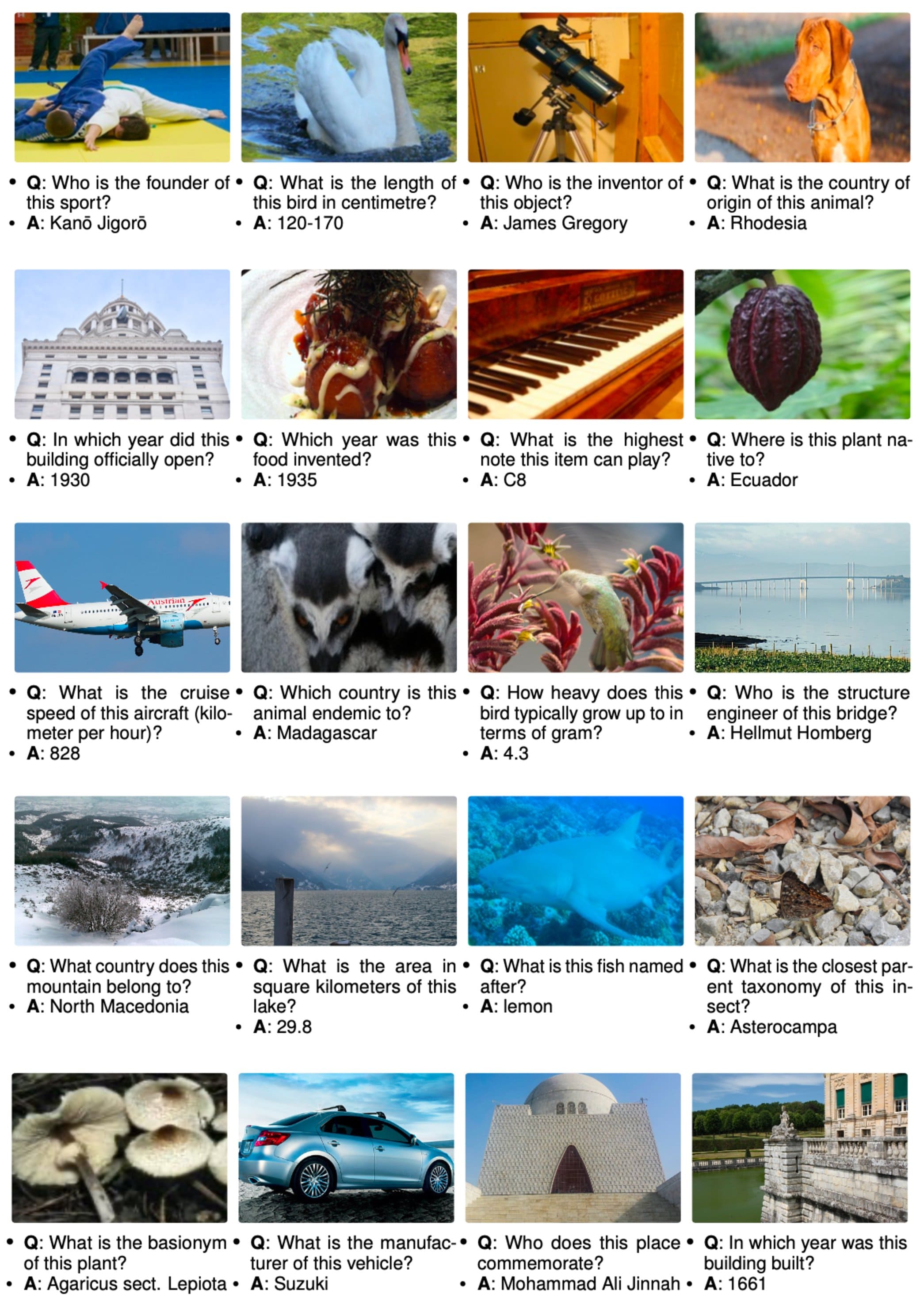}
    \caption{Random examples from the training set of \infoseek$_\text{Wikidata}$.}
    \label{fig:question_prefix}
\end{figure*}

\begin{figure*}[!ht]
    \centering
    \includegraphics[width=\textwidth]{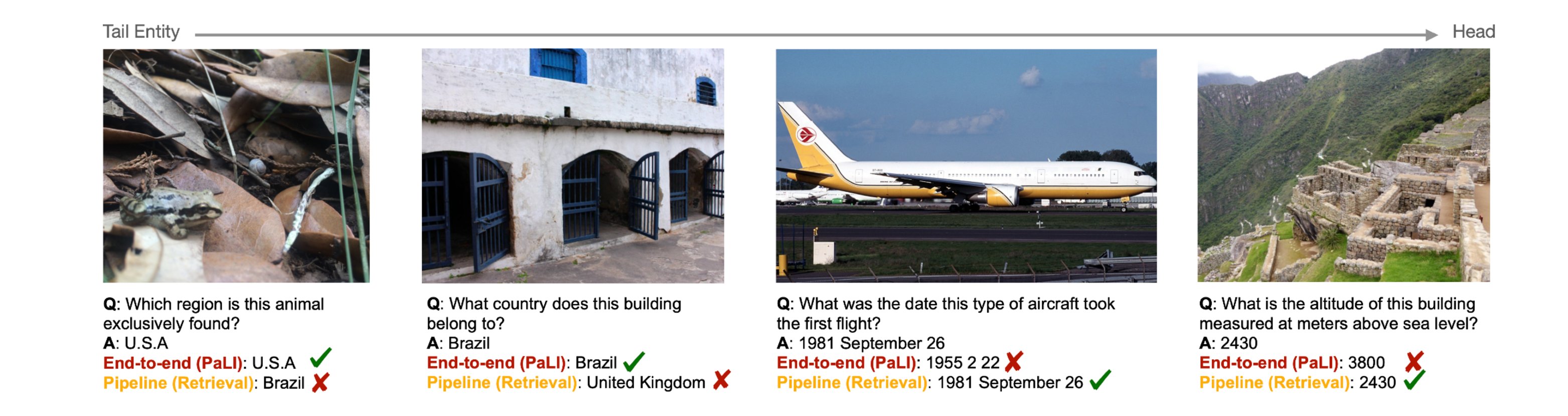}
    \vspace{-6mm}
    \caption{\textbf{Examples} of predictions of PaLI-17B and CLIP$\to$FID on \ourdataset (left to right shows tail to head entities).
    }
    
    \label{fig:error_analysis}
\end{figure*}

\begin{figure*}[ht!]
    \centering
    \includegraphics[width=\textwidth]{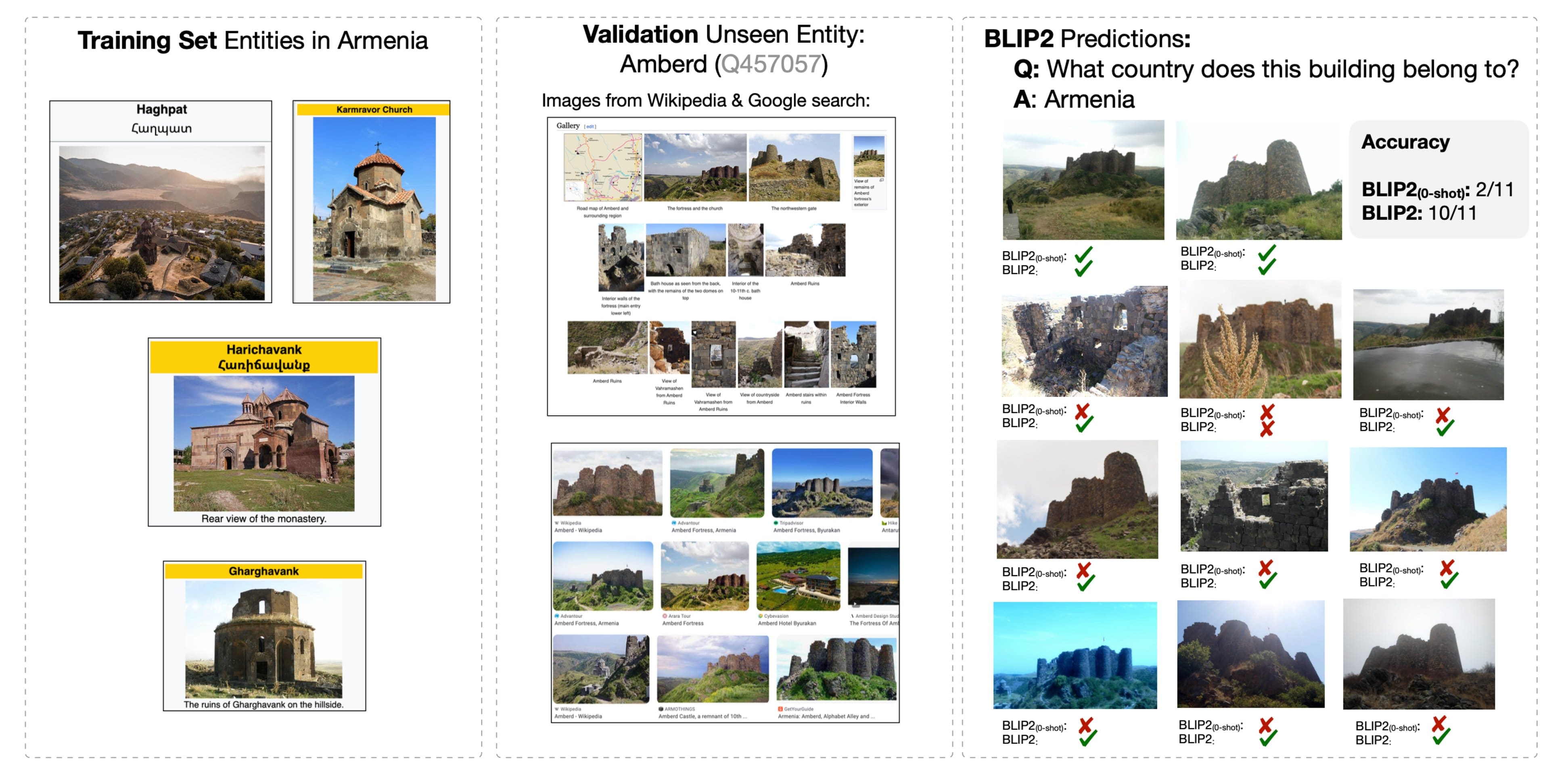}
    \caption{\textbf{Examples} of predictions of BLIP2$_{\texttt{(0-shot)}}$ and BLIP2 on \ourdataset (Entity: Q457057). Fine-tuning improves the accuracy from 2/11 to 10/11, despite it being an \textsc{unseen entity} (not in the training set). We show training set entities that are located in Armenia and images of Amberd on the internet.}
    \label{fig:blip2_generalization}
\end{figure*}

\begin{figure*}[ht!]
    \centering
    \includegraphics[width=0.45\textwidth]{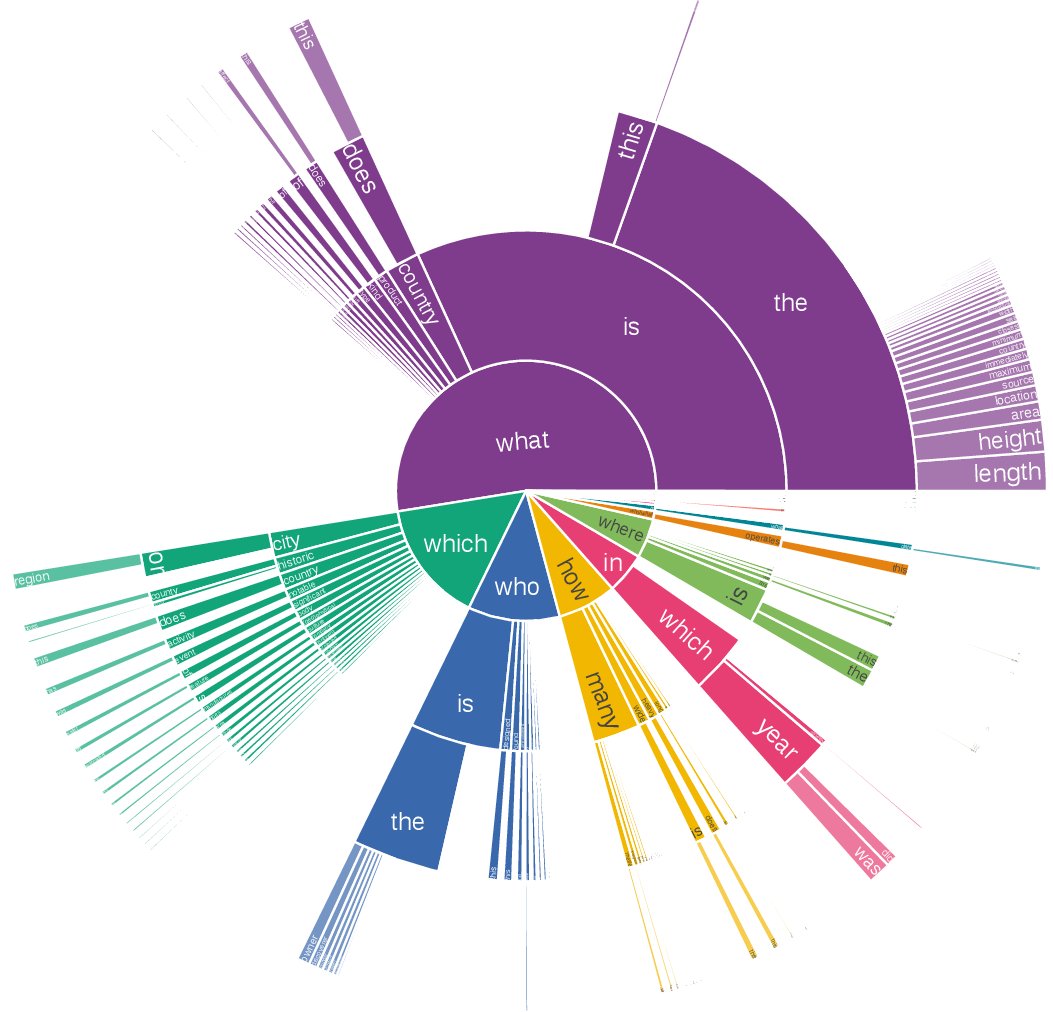}
    \includegraphics[width=0.45\textwidth]{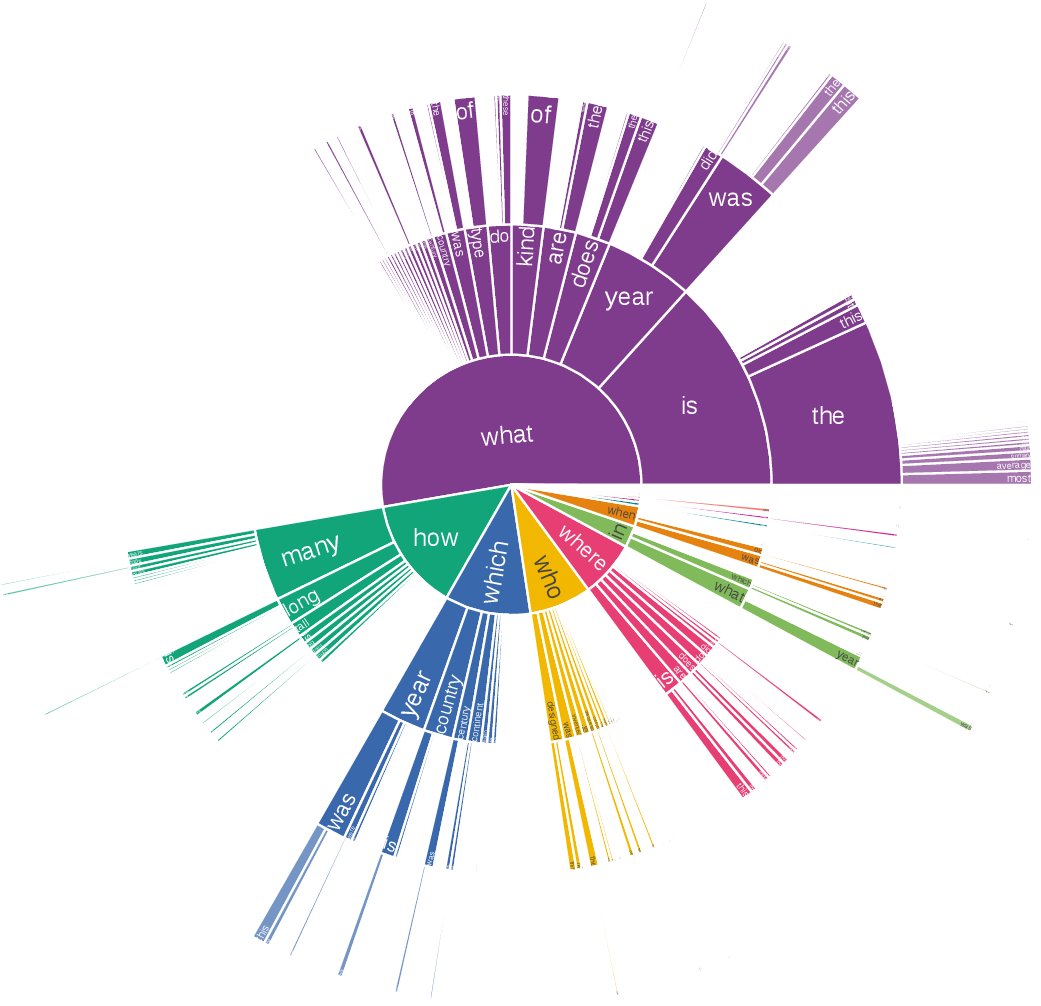}
    \caption{Question prefix distribution in \infoseek$_\text{Wikidata}$ (left) and \infoseek$_\text{Human}$ (right).}
    \label{fig:question_prefix}
\end{figure*}

\begin{table}
\centering
\tiny
\begin{tabular}{lrrrr}
\toprule
  &  \#\textsc{Unseen} & \multirow{2}{*}{\#Total} &  Question Type & \#Entity\\
    &  \textsc{Question/Entity} & & \textsc{Time/Num./Str.} &\\
    \midrule
    Train & - / - & 934,048 & 4.4/20.4/ 75.2\% & 5,549\\
    Val & 18,656/54,964	& 73,620	&4.6/	21.6/	73.8\%&		1,794\\
    Test & 98,901/249,079	& 347,980	&4.8/22.9/72.3\%	&	8,905\\
    Human & 3,248/5,683&	8,931&	26.8/	26.4/46.8\%&		806\\
\bottomrule
\end{tabular}
\caption{\textbf{\ourdataset Dataset statistics.} Average question per image rate is 1.4 and 1.0 for Wikidata and Human split, respectively.}
\label{tab:opera_stat}
\end{table}

\begin{figure}[ht!]
    \centering
    \includegraphics[width=0.45\textwidth]{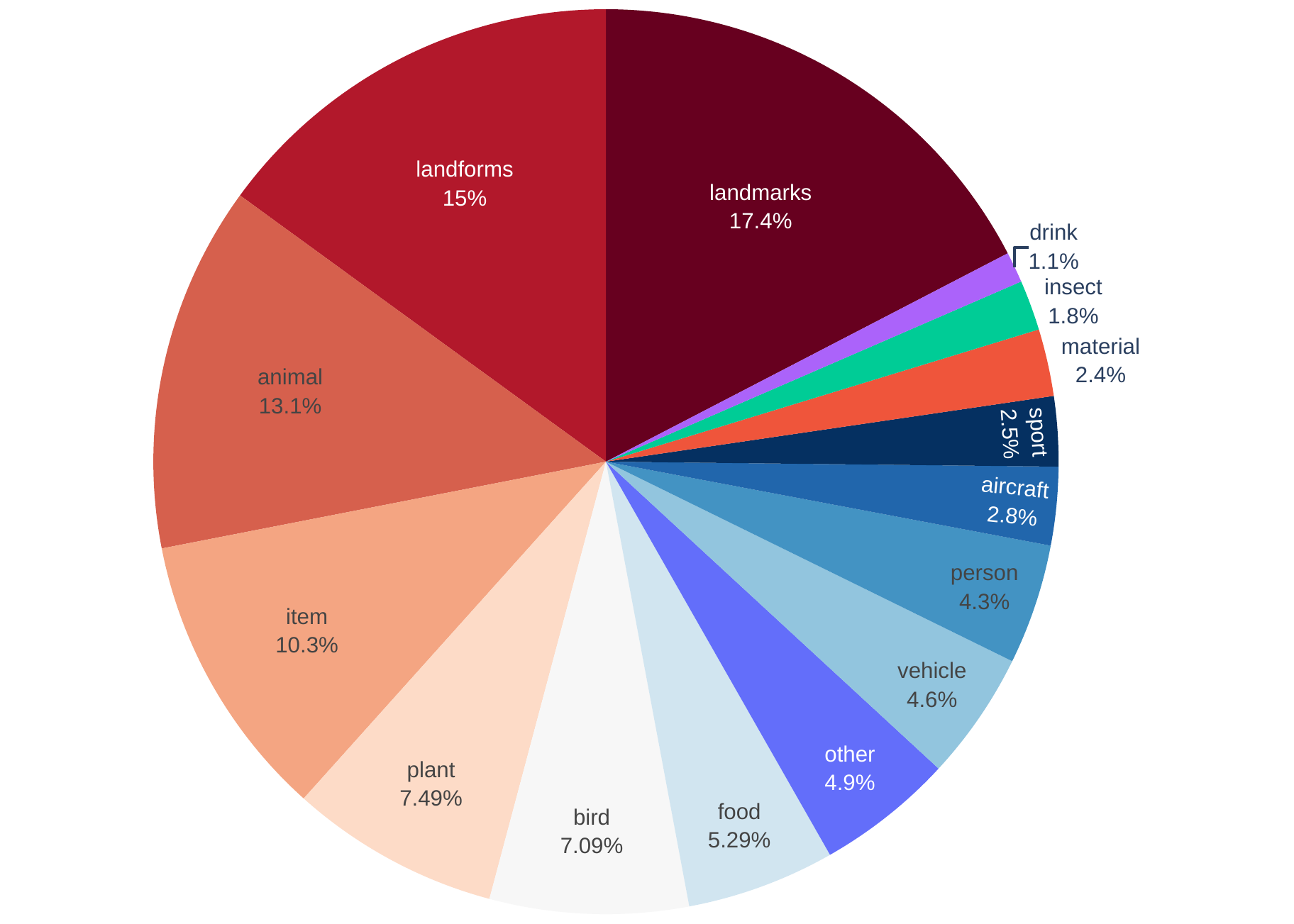}
    \caption{Distribution of the entities in \ourdataset (Grouped by their super category).}
    \label{fig:entity_distribution}
\end{figure}

\section{Details of the Dataset.}
In this section, we provide more details of the human annotation quality control and automatic data generation process. 
We summarize the statistics of \ourdataset in Table~\ref{tab:opera_stat} and show question prefix distribution in Figure~\ref{fig:question_prefix} and entity distribution in Figure~\ref{fig:entity_distribution}.

\subsection{Human Annotation Quality Control}
\begin{figure*}[ht!]
    \centering
    \includegraphics[width=\textwidth]{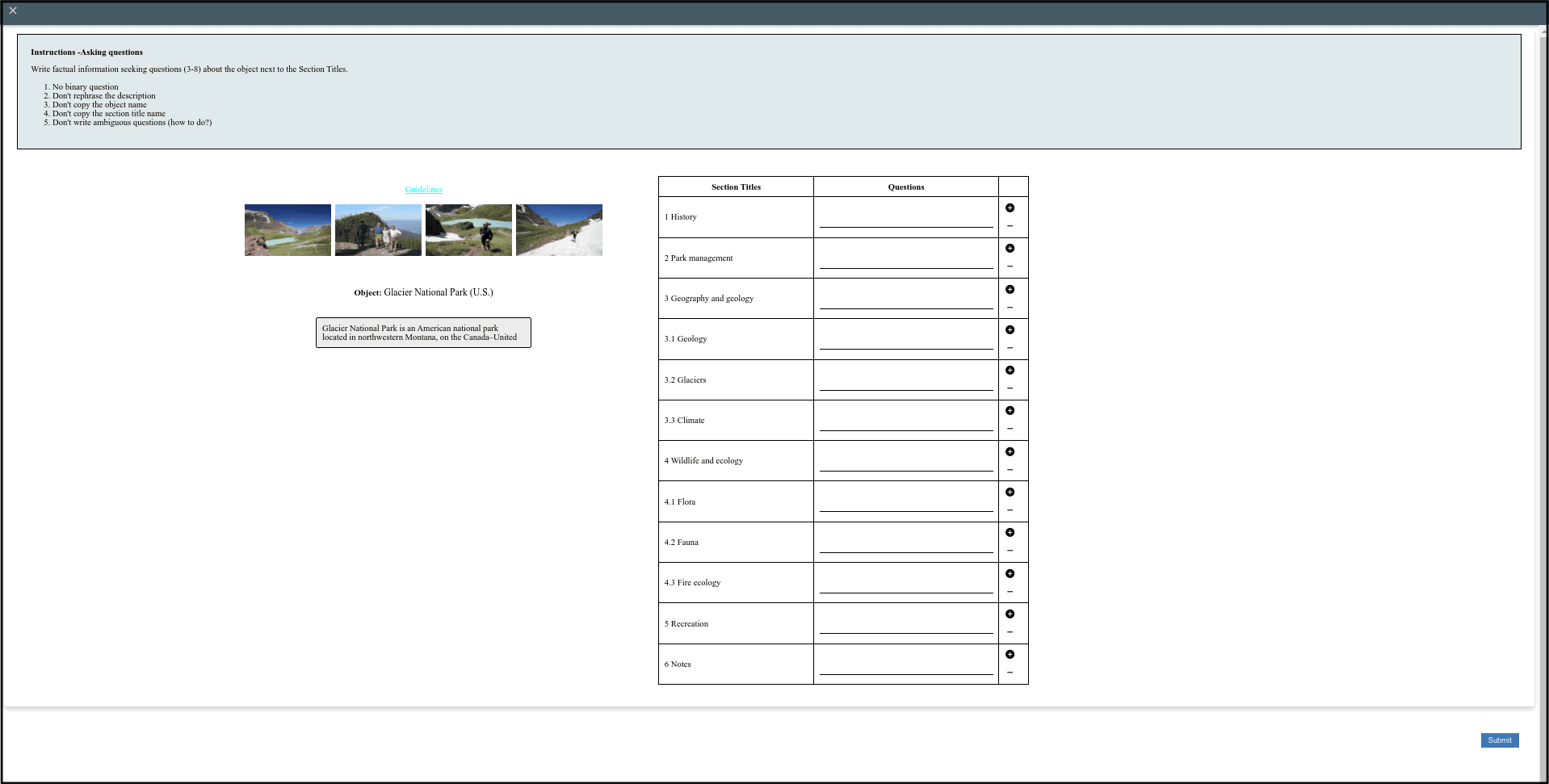}
    \includegraphics[width=\textwidth]{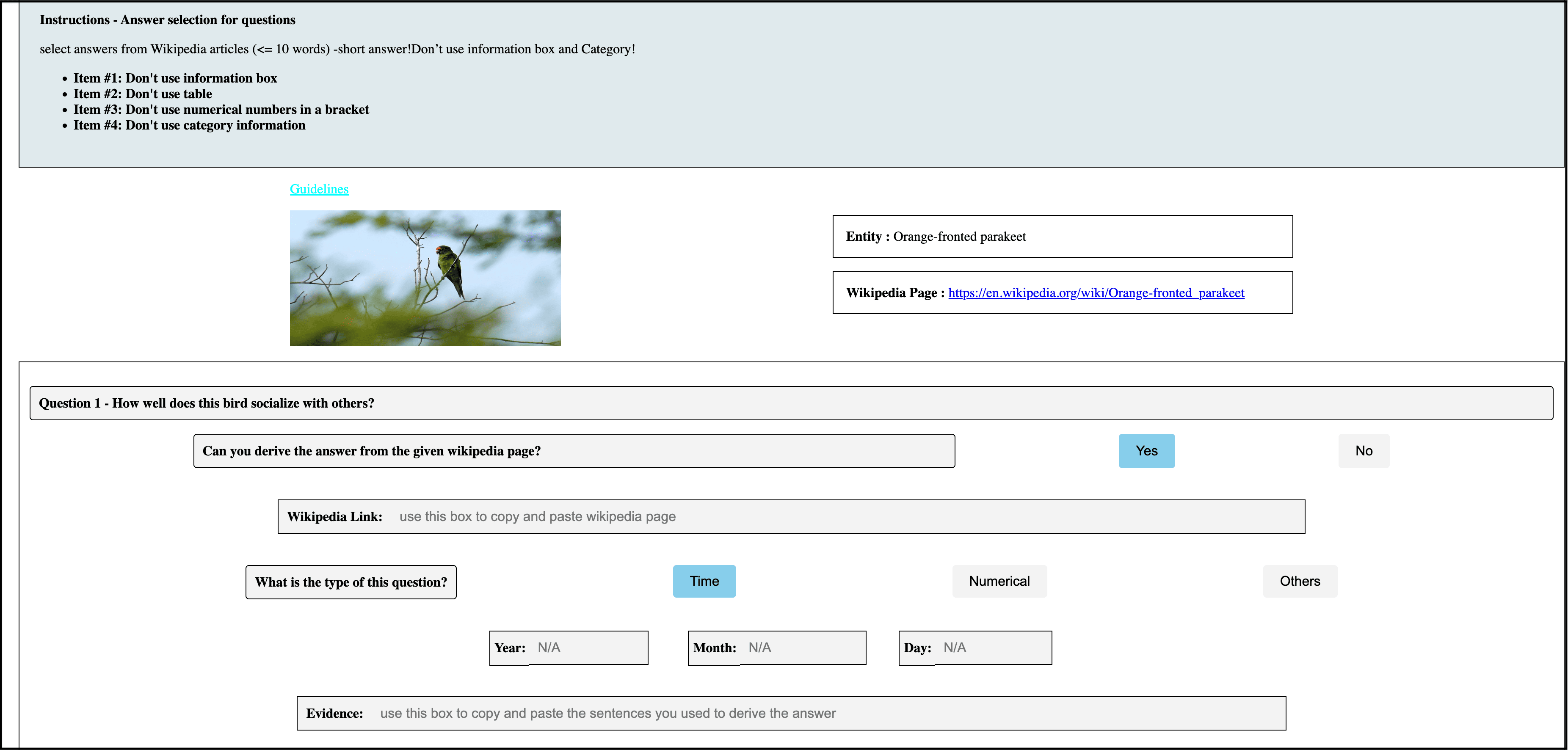}
    \caption{Annotation Interface for Stage 1 (Top) and Stage 2 (Bottom).}
    \label{fig:annotation_interface}
\end{figure*}

\custompara{Instruction and Training.} 
We hire 30 full-time in-house annotators to collect questions and answers in $\ourdataset_{\texttt{Human}}$. Annotators are native English speakers in the U.S. and are aware of the purpose of the collected data.
To ensure the quality of annotations in the $\ourdataset_{\texttt{Human}}$ dataset, a comprehensive training process was designed and implemented for our annotators. This process involved a pilot study, in which annotators read the instructions and annotated a few sample examples, followed by a tutorial session and a quiz. The tutorial was conducted through an online video session and provided a comprehensive overview of the instructions while addressing common mistakes identified in the pilot study. Only annotators who passed the quiz were selected to work on the main task, with 30 annotators completing the training. 
We hire annotators at \$17.8 per hour, which is higher than the minimum wage in the U.S., to fairly compensate annotators for their time and effort. The average completion time for stages one and two of the annotation task was 12 and 10 minutes, respectively. A screenshot of the annotation interface is provided in Figure~\ref{fig:annotation_interface}.

\custompara{Annotation Procedure}\\
\textit{Stage 1 (Question Writing)}: As shown in Figure~\ref{fig:annotation_interface} (Top), annotators are shown with images of a visual entity on the left-hand side with a short description of the entity from Wikipedia below. On the right, we show a list of Wikipedia section titles of the entity and ask annotators to write relevant questions next to the section title. We prevent annotators from asking binary questions, asking visual attributes (such as color), writing questions by rephrasing the description, copying entity names and section titles into the question, and avoiding writing ambiguous questions.
\\\\
\textit{Stage 2 (Answer Labeling)}: As shown at the bottom of Figure~\ref{fig:annotation_interface}, annotators are present with info-seeking questions to the entity collected from Stage 1 and a Wikipedia link of the entity. For each question, annotators are asked to find a short span of answers (less than 10 words) from the Wikipedia page. They are asked to answer two questions: (1) ``Can you derive the answer from the given Wikipedia page?'' and (2) ``What is the type of this question?'' and select from three options (\textsc{Time}, \textsc{Numerical}, \textsc{Others}).
For each answer, they will then fill in the answer box (\textsc{Time}: \snlp{[year, month, day]}, \textsc{Numerical}: \snlp{[min, max, unit]}, \textsc{Others}: \snlp{[string]}) and copy paste a short sentence from Wikipedia that contains the answer to the evidence section.
We decided to exclude questions without answer spans from Wikipedia following TyDiQA-GoldP as the dataset is already hard enough and reserve these questions for future work.

\begin{table}
\centering
\small
\begin{tabular}{ccc}
\toprule
& Correct & Incorrect \\
\midrule
Percentage & 95\% & 5\% \\
\bottomrule
\end{tabular}
\caption{Expert judgments of answer accuracy based on a sample of 200 examples from \infoseek$_{\text{Human}}$.}
\label{tab:expert_feedback}
\vspace{-5mm}
\end{table}

\custompara{Expert Feedback and Correction.} 
Expert annotators provided regular feedback during annotation and conducted thorough post-annotation verification. The data was split into three batches, with annotators flagged and provided feedback for those who consistently made similar mistakes. After the completion of stage 1, questions that revealed the entity name, asked about the color or shape of an object, or were binary were automatically rejected. After stage 2, three expert annotators reviewed and processed the question-answer pairs, removing unqualified pairs and verifying the answer span from the annotated evidence sentence. Rejected pairs may have included questions that were not answered by the annotated answer or were too general and resulted in an ambiguous answer. The expert annotators also corrected the question type annotation and edited the answer span into the correct format, such as adding units for numerical questions or shortening long answer spans that exceeded ten tokens. 
Finally, the expert annotators reviewed the image-question-answer triples to reject bad images or clarify the question when multiple objects were present in the image. For example, a building was specified when multiple buildings were present in the image. On average, it took 1.5 hours to verify 1000 triples, as the majority of images contained a single object.

Following TyDiQA~\citep{clark2020tydi}, we analyze the degree to which the annotations are correct instead of the inter-annotator agreement since the question may have multiple correct answers. In Table~\ref{tab:expert_feedback}, human experts carefully judged a sample of 200 examples from \infoseek$_\text{Human}$ split. For each example, the expert reads through the Wikipedia page of the queried visual entity and finds the answer to the question. They then indicate whether the annotated answer is correct.
We take the high accuracy (95\%) as evidence that the quality of the dataset offers a valuable and reliable signal for evaluating visual info-seeking models.

\subsection{Filtering and Subsampling}
\label{appendix:verify_sample}
\custompara{Filtering.}
To test the models' ability to answer visual information-seeking questions that require fine-grained knowledge, which can be learned from the pre-training corpus such as Wikipedia, we need to verify the consistency of answers between Wikidata and Wikipedia. Given that Wikidata and Wikipedia are crowd-sourced independently, some QA pairs created from Wikidata may not be present in the Wikipedia article or may have different answers. Therefore, we filtered out QA pairs where the answer could not be found in the Wikipedia article of the entity. We performed an exact string match to verify answers for string questions and used fuzzy string matching~\footnote{\texttt{SequenceMatcher} from \texttt{difflib} library} with a substring ratio greater than 0.9 if an exact match could not be found. For time questions, we applied an exact match to verify the year, month, and date. In some cases, the year of construction of a building varied by a year, so we allowed a +/- 1 year deviation for the time question. For numerical questions, we used exact matching to verify the numbers in the article. However, in many cases, the units were different (meters or inches), or a range with a minimum and maximum was given. 
We used regular expressions to extract the number or range from the Wikipedia article and filter out the QA pairs if it is counted as incorrect based on the ``Relaxed accuracy'' in Section 3.3 in the main text. 
Based on a manual analysis of 200 randomly sampled QA pairs, we found that 97\% of the answers of the \ourdataset$_\text{Wikidata}$ could be found in the Wikipedia article.

\custompara{Subsampling Questions.}
In order to achieve a more diverse set of questions in \ourdataset, we applied a subsampling method to address the skewed distribution of crowd-sourced knowledge triples in Wikidata. The method followed the approach used in \citet{zhong2022romqa}. This involved defining $P(r,c)$ as the percentage of triples that contain the relation $r$ and the subject entity's category as $c$. The $P'(r,c)=1/|(r,c)|$ was calculated as the average probability of a relation-category pair and Image-Q-A triples were removed with increasing likelihood based on the probability $r=1 - min(1, P(r,c)'/P(r,c))^{1/2}$. Additionally, the same subsampling method was applied to balance the answer distribution for each relation. This resulted in the question prior baseline achieving a relatively low score (3.2) in \ourdataset$_\text{Wikidata}$, as shown in Table 4 in the main text.

\subsection{Evaluation Metric.}
\label{appendix:metric}
There are three types of questions, \ie, \textsc{String}, \textsc{Time}, and \textsc{Numerical}, which are evaluated differently. 
Particularly, we adopt the VQA accuracy~\cite{balanced_vqa_v2,marino2019ok} against multiple references for \textsc{String} and \textsc{Time} questions, and utilize {\em Relaxed Accuracy}~\cite{methani2020plotqa,masry2022chartqa} for \textsc{Numerical} questions.
For \textsc{String} questions, we use the alias of answers from Wikidata as multiple references for \ourdataset$_\text{Wikidata}$ (\#avg = 4.5), and the human-annotated multiple references for \ourdataset$_\text{Human}$ (\#avg = 2.4).
\noindent\fbox{%
\small
    \parbox{\linewidth}{%
        Exact Match: Correct if the prediction matches any one of the references exactly.\\
        \\
        $\bullet$	 prediction=``USA", references=[``USA", ``U.S.", ``United States of America", ...] $\to$ \checkmark
    }%
}

For \textsc{Time} questions, the answer references account for different date formats of year/month/day.
Meanwhile, we perform a relaxed match (with a one-year error tolerance) to measure the model's prediction, because it is quite often that historical events are only associated with estimated time.
\noindent\fbox{%
\small
    \parbox{\linewidth}{%
        Exact Match: Correct if the prediction matches any one of the references exactly.\\
        \\
        $\bullet$	 prediction=``1991", references=[``1990", ``1991", ``1992"]) $\to$ \checkmark \\
        $\bullet$	 prediction=``1991 6 11", references=[``1991 6 11", ``1991 June 11", ``11 June 1991", ...] $\to$ \checkmark
    }%
}

For \textsc{Numerical} questions, the exact match would not be able to handle the case where a range (\eg, a pair of minimum and maximum values) is provided as annotated ground truth. To account for this, we make a slight modification to the Relaxed Accuracy with a 10\% tolerance range. 
\noindent\fbox{%
\small
    \parbox{\linewidth}{%
        Relaxed Accuracy: correct if the prediction is in the reference range or the prediction range overlaps with the reference range of more than 50\%\\
        1) ref\_min $\leq$ pred $\leq$ ref\_max;\\
        2) IoU([pred\_min, pred\_max], [ref\_min, ref\_max]) $\geq$ 50\%\\
        \\
        $\bullet$	reference= 10 cm $\to$ reference= [9, 11] \texttt{\# 10\% tolerance} \\
        $\bullet$	prediction= 10, reference= [9, 11] $\to$ \checkmark\\
        $\bullet$	prediction= [5, 6], reference= [9, 11] $\to \times$
    }%
}
A single value prediction is counted as correct if it falls within the answer range, and a range prediction is correct if the intersection-of-union between the prediction and answer is greater than or equal to 50\%.
Finally, we calculate the accuracy for each data split (\textsc{Unseen Question} and \textsc{Unseen Entity}), and take the harmonic mean of them as the overall accuracy.

\subsection{Image Sources.}
Image Recognition (or Retrieval) Datasets: ImageNet21k-P ~\cite{russakovsky2015imagenet,ridnik2021imagenet}, iNaturalist2017~\cite{van2018inaturalist}, Cars196 ~\cite{krause20133d}, SUN397~\cite{xiao2010sun}, Food101~\cite{bossard2014food}, 
Sports100 ~\cite{sport100}, Aircraft~\cite{maji2013fine},
Oxford Flower~\cite{nilsback2008automated}, Google Landmarks v2~\cite{weyand2020google}.

\section{Implementation details of the baseline systems}
In this section, we provide complete implementation details of baseline models for the \ourdataset task. We summarize hyperparameters for fine-tuning in Table~\ref{tab:hpyer}.

\subsection{without-KB Models}
\custompara{\palift and PaLI-X.}
We fine-tuned a 17B PaLI~\cite{chen2022pali} and 55B PaLI-X~\cite{chen2023pali} on \ourdataset training set using the \nlp{``answer in en: [question] <extra\_id\_0>''} prompt.

\custompara{BLIP2 and InstructBLIP.}
We fine-tuned a BLIP2~\cite{li2023blip} and InstructBLIP~\cite{dai2023instructblip} on \ourdataset training set using the \nlp{``Question: [question] Short answer:''} prompt with the \texttt{LAVIS} library~\citep{li-etal-2023-lavis}. The length penalty is set to -1.
Since BLIP2 models present zero-shot capabilities on \ourdataset, we employ early stopping to prevent over-fitting on the training set based on the performance on the validation set.

\custompara{\ofaft.}
We fine-tuned the OFA$_\text{large}$~\cite{lu2022unified} model for 20k steps. During inference, we apply beam search decoding with a beam size set to 5. OFA achieves 11.7 and 4.0 on \ourdataset Wikidata and Human split, respectively.

\custompara{mPLUG-owl.}
We fine-tuned mPLUG-owl~\citep{ye2023mplug} for 10k steps with a learning rate of 2e-4 and batch size of 1 using LoRA~\citep{hu2021lora}. mPLUG-owl achieves 7.7 on \ourdataset Human split.

\custompara{PaLM(Q-Only).}
We use PaLM 540B~\cite{chowdhery2022palm} in-context learning under the 5-shot setting with the following prompt:

\noindent\fbox{%
\small
    \parbox{\linewidth}{%
        Please answer the following question.

 question: \{\texttt{Question\_1}\}. answer: \{
 \texttt{Answer\_1}\}. \\
...\\
 question: \{\texttt{Question\_i}\}. answer: 
    }%
}

\subsection{With-KB Models}

\begin{table}[ht!]
\centering
\tabcolsep 2pt
\tiny
\begin{tabular}{lccccc}
\toprule
 & \palift & PaLI-X & (Instruct)BLIP2 & \ofaft & FID \\
\midrule
Optimizer & Adafactor & Adafactor & Adam & Adam & Adafactor \\
Batch size & 128 & 128 & 16 & 512 & 64\\
Train steps & 10k & 800 & 400 & 20k & 200\\
LR & 1e-4 & 1e-4 & 5e-5 & 5e-5 & 2e-4\\
LR scheduler & linear decay & constant & constant & polynomial decay & constant \\
Warmup steps & 1000 & 1000 & - & 1000 & - \\
Image size & 224 & 224 & 224 & 480 & - \\
Beam size & 5 & 5 & 5 & 5 & 5\\
Vision backbone & ViT-e & ViT-22B & ViT-g & ResNet152 & -\\
LM backbone & mT5$_\text{XXL}$ & UL2-32B & Flan-T5$_\text{XXL}$ & BART$_\text{large}$ & T5$_\text{large}$\\
\#Params & 17B & 55B & 12.1B & 0.4B & 0.4B\\
Computing & 32 TPU$_{\text{v4}}$ & 64 TPU$_{\text{v4}}$ & A40 & 8 A100 & 64 TPU$_{\text{v4}}$\\
Time & 6 hours & 1 hour & 1 hour & 48 hours & 1 hour\\
\bottomrule
\end{tabular}
\caption{Hyperparameters for fine-tuning models on \ourdataset.}
\label{tab:hpyer}
\end{table}

\custompara{PaLM.}
We use PaLM 540B~\cite{chowdhery2022palm} in-context learning under the 5-shot setting with the prompt present below. The \texttt{Entity\_1} is the gold entity provided in the training set (with KB setting). The \texttt{Entity\_i} is the top-1 prediction from the entity linking stage of the queried image.

\noindent\fbox{%
\small
    \parbox{\linewidth}{%
        Please answer the following question.

 question: \{This is \texttt{Entity\_1}. \texttt{Question\_1}\}. answer: \{
 \texttt{Answer\_1}\}. \\
...\\
 question: \{This is \texttt{Entity\_i}. \texttt{Question\_i}\}. answer: \{
    }%
}

\custompara{FID.}
The T5$_\text{large}$ FID~\cite{izacard2020fid} model was fine-tuned in two stages using 100 passages with a maximum input length of 192 tokens.
To form synthetic training data with (passage, question, answer) triples, we combine oracle entity passage with linked entity (from EntLinker) passages.
We fine-tune the model on Natural Questions~\cite{kwiatkowski-etal-2019-natural} for 10k steps and then continue to fine-tune it on \ourdataset for 200 steps with a batch size of 64.
\noindent\fbox{%
\small
    \parbox{\linewidth}{%
 question: This is \texttt{Entity}. \texttt{Question}. context:
 \texttt{Passage}}%
}

\section{Additional Experiment Results}
\begin{table*}[!ht]
\centering
\small
\begin{tabular}{@{\;}lcc@{\;}cccccc@{\;}}
\toprule
\multirow{3}{*}{\textbf{Model}} & \multirow{3}{*}{\textbf{\# Params}}&  \multirow{3}{*}{\textbf{Components use KB}} & \multicolumn{3}{c}{$\textbf{\ourdataset}_{\text{Wikidata}}$} & \multicolumn{3}{c}{$\textbf{\ourdataset}_{\text{Human}}$}\\
\cmidrule(lr){4-6}\cmidrule(lr){7-9} & & & \small \textsc{Unseen}  & \small \textsc{Unseen} & \multirow{2}{*}{\small Overall} & \small \textsc{Unseen} & \small \textsc{Unseen}  & \multirow{2}{*}{\small Overall} \\
& & & \small \textsc{Question} & \small  \textsc{Entity} & & \small \textsc{Question} &  \small \textsc{Entity} &  \\
\midrule
CLIP $\to$ PaLM & 540B & CLIP & 21.9	& 18.6	& 20.1& 15.6	& 14.9 &	15.2 \\
CLIP $\to$ FiD & \pz\pz1B & CLIP \& FiD & 20.7	& 18.1&	19.3 & 18.9 & 	17.6	& 18.2 \\
\bottomrule
\end{tabular}
\caption{
\ourdataset full results on \withkb setting.
}
\label{tab:withkb}
\end{table*}
\custompara{Complete numbers for With-KB Models.}
We show the complete results for With-KB models in Table~\ref{tab:withkb}.

\begin{table}[!tbh]
\small
\centering
\tabcolsep 2.5pt
\begin{tabular}{lccc}
\toprule
\multirow{3}{*}{\textbf{Model}} & \multicolumn{3}{c}{$\textbf{\ourdataset}_{\text{Wikidata}}$}\\
\cmidrule(lr){2-4}
  & \small \textsc{Unseen}  & \small \textsc{Unseen} & \multirow{2}{*}{\small Overall} \\
  & \small \textsc{Question} &\small  \textsc{Entity} & \\
\midrule
{\textbf{Without-KB Setting}}\\
Prior &  \pz4.6	& \pz2.5 & \pz3.2 \\
\qonly &  \pz5.5 & \pz4.2 & \pz4.8 \\
InstructBLIP & 15.0 & 14.0& 14.5\\
BLIP2 & 15.0& 14.2& 14.6\\
\palift-17B  & 24.2 &	16.7	& 19.7\\
PaLI-X & 25.8 & 22.4 & 24.0\\
\midrule
{\textbf{With-KB Setting}}\\
CLIP $\to$ PaLM & 22.7	& 18.5	& 20.4\\
CLIP $\to$ FiD &   23.3 & 19.1 & 20.9\\
\rowcolor{lightgray} Oracle $\to$ FiD & 52.1 & 53.0 & 52.5\\
\bottomrule
\end{tabular}

\vspace{-2mm}
\caption{\ourdataset full results on Wikidata \textbf{validation} set.
}
\label{tab:val_result}
\end{table}

\begin{table}[!tb]
\centering
\small
\tabcolsep 5pt
\begin{tabular}{lccc}
\toprule
\multirow{2}{*}{\textbf{Model}}&  \small \textsc{Time} & \small \textsc{Numerical} & \small \textsc{String} \\
 & \scriptsize (Acc.) & \scriptsize (Relaxed Acc.) & \scriptsize (Acc.)\\
\midrule
\textbf{\nokb Setting}\\
Prior & \pz\pz0	& \pz3.5	& \pz2.3 \\
\qonly & \pz\pz4.6	&11.0&	\pz2.7 \\
InstructBLIP & \pz6.6 & \pz8.2 & 16.1 \\
BLIP2 & \pz5.6 & \pz6.0 & 17.0 \\
\palift-17B & \pz1.0	& 14.8	& 18.2\\
PaLI-X & \pz8.1	& 17.2	&24.8\\
\midrule
\textbf{\withkb Setting}\\
CLIP $\to$ PaLM & 17.8	& 21.3	&17.7\\
CLIP $\to$ FiD & 13.8	& 15.2	& 20.5\\
\bottomrule
\end{tabular}
\vspace{-2mm}
\caption{\textbf{Results w.r.t. each question types} on the ${\ourdataset}_{\text{Wikidata}}$ val set of unseen entity split.
}
 \vspace{-4mm}
\label{tab:question_type_entity}
\end{table}
\custompara{Complete numbers for \ourdataset$_{\text{Wikidata}}$ Validation set.}
We show the complete Validation results for Without-KB and With-KB models in Table~\ref{tab:val_result} and question type score of unseen entity split in Table~\ref{tab:question_type_entity}.

\subsection{OK-VQA Annotation Guidelines}
\label{appendix:okvqa}
Five adult annotators each annotate 100 examples  (500 in total) sampled from the OK-VQA training set. Annotators are instructed to categorize each example into one of three categories (see Table~\ref{tab:okvqa}).

\begin{table}[h!]
\centering
\scriptsize
\begin{tabular}{lc}
\toprule
\textbf{Question Category} & \textbf{Percentage} \\
\midrule
Answered directly by looking at the corresponding image & 50.8\% \\
Answered without looking at the image (Q-only) & 20\% \\
Requiring a Google search for an answer & 29.2\% \\
\bottomrule
\end{tabular}
\caption{OK-VQA annotation results.}
\label{tab:okvqa}
\end{table}

\label{sec:appendix}

\end{document}